%% file: acl_latex_arxiv_submission_v1.tex
\pdfoutput=1

\documentclass[11pt]{article}

\usepackage[final]{acl}

\usepackage{times}
\usepackage{latexsym}

\usepackage[T1]{fontenc}

\usepackage[utf8]{inputenc}

\usepackage{microtype}

\usepackage{inconsolata}

\usepackage{graphicx}

\usepackage{arydshln}
\usepackage{ulem}
\usepackage{booktabs}
\usepackage{listings}
\usepackage{multirow}
\usepackage{pifont}
\usepackage{comment}
\usepackage{amsmath}
\usepackage{subcaption}

\input{latex/listing_styles}

\definecolor{gold}{rgb}{0.6, 0.4, 0}
\definecolor{darkgreen}{rgb}{0.0, 0.5, 0.0}
\newcommand{\colordarkgreen}[1]{\textcolor{darkgreen}{#1}}
\newcommand{\paragraphsquish}[1]{\noindent\textbf{#1}}

\newcommand{\datasetname}[0]{Dialogue-SWEBench}

\newcommand{\bdk}[1]{\textcolor{purple}{\bf\small [#1 --bking2]}} %
\newcommand{\status}[1]{\textcolor{gold}{\bf\small Status: #1}} %
\newcommand{\jmf}[1]{\textcolor{blue}{\bf\small [#1 --JMF]}} %
\renewcommand{\bdk}[1]{}
\renewcommand{\jmf}[1]{}
\renewcommand{\status}[1]{}

\input{latex/notation}

\title{\datasetname: A Benchmark for Dialogue-Driven Coding Agents}

\author{Brendan King \and Jeffrey Flanigan\\
  University of California, Santa Cruz \\
  \texttt{\{bking2,jmflanig\}@ucsc.edu}}

\begin{document}
\maketitle

\begin{abstract}
AI coding agents have rapidly transformed software engineering, powering widely used interactive coding assistants.
Despite their interactive real-world use, existing benchmarks evaluate them as fully-autonomous systems.
In this work, we introduce \textbf{\datasetname}, an automatic benchmark dataset for evaluating the ability of coding agents to resolve real-world software engineering problems through dialogue with a user.
We design a novel, persona-grounded user simulator to support our task evaluation, and augment our task evaluation with automatic evaluations of dialogue quality.
We also propose a new \textit{schema-guided} agent, aimed at improving the dialogue capabilities of off-the-shelf coding agents, which improves over strong baselines by 3-14\%.
Our results indicate that better coding models do not always correspond to better dialogue models, suggesting that dialogue capability is a distinct and currently understudied dimension of coding agent performance.\footnote{Code and data available at \href{https://jlab-nlp.github.io/dialogue-swe-bench/}{https://jlab-nlp.github.io/dialogue-swe-bench/}}
\end{abstract}

\section{Introduction}
\label{sec:introduction}

Coding agents have radically transformed the software-engineering (SWE) landscape, powering widely-used tools like Github Co-Pilot and Claude Code \cite{github2022copilot, anthropic2025claudecode}.
To support their development, the research community has produced increasingly complex benchmarks for evaluating coding agents on real-world engineering tasks \cite{jimenez_swe-bench_2024, Merrill2026TerminalBenchBA}.
This has enabled rapid improvement in the \textit{coding capabilities} of these agents, through advances in model reasoning \cite{openai_gpt-5_2025, qwen3technicalreport, anthropic2025claude4}, coding agent design \cite{yang_swe-agent_2024, wang_openhands_2024}, and 
training methods for agentic-coding \cite{pan_training_2024, wei_swe-rl_2025}.

Yet despite this progress, existing benchmark datasets have focused on the fully autonomous setting, leaving the interactive nature of real-world coding agent use understudied. Such fully-autonomous SWE evaluations leave a significant gap between what they measure and real-world software engineering.
First, fully-autonomous benchmarks presume a complete and correct problem specification as input, yet these are rarely available in practice.\footnote{For example, \citet{chowdhury_introducing_2024} find that 76\% of the real-world Github Issues comprising SWE-Bench \cite{jimenez_swe-bench_2024} are at least somewhat under-specified, and 39\% are deemed too vague to determine ``what a successful solution would look like.''}
Second, real-world engineering with coding agents is highly interactive: in a study of real-world agentic coding sessions, \citet{baumann_swe-chat_2026} find that users use dialogue to correct or reject agent outputs 44\% of the time.
This interactivity was also largely one-sided: while users frequently used dialogue to correct agent outputs, agents themselves only sought clarification 1-2\% of the time.
These findings suggest that improving the \textit{dialogue capabilities} of coding agents represents a significant and largely untapped opportunity for real-world impact.

\begin{figure}
    \centering
    \includegraphics[width=\columnwidth]{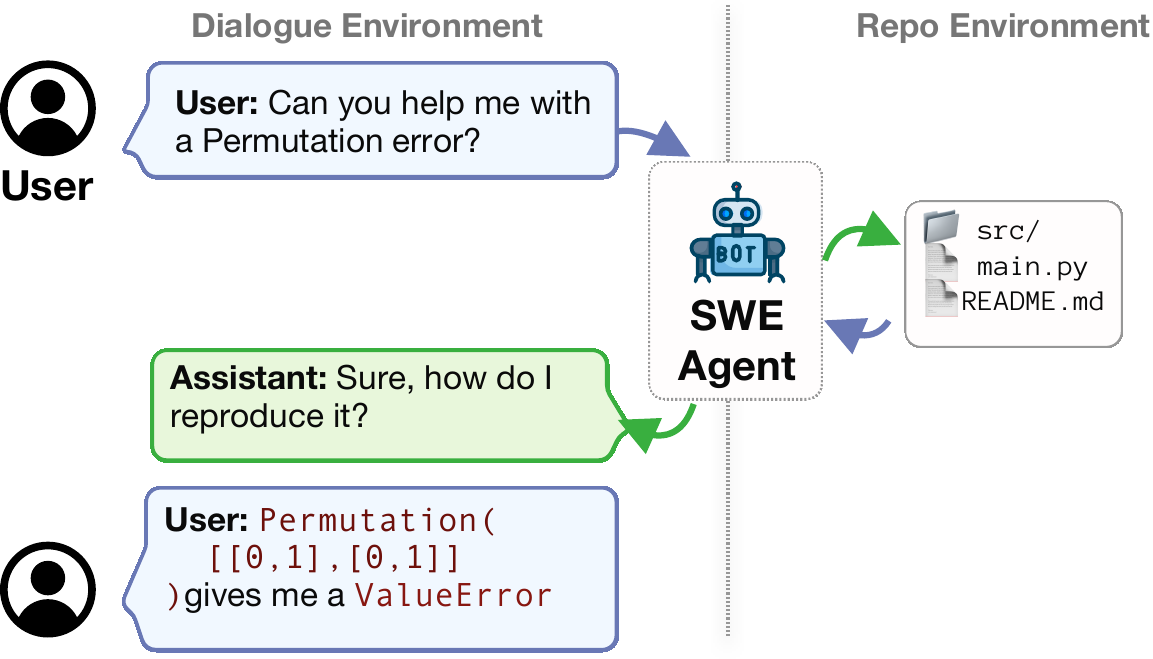}
    \caption{Our dialogue benchmark for coding agents. The user never interacts with the code, and performs software engineering tasks strictly through dialogue. Agents complete repository-level software engineering tasks through dialogue with the user.}
    \label{fig:intro-1}
\end{figure}

\begin{figure*}[h!]
    \centering
    \includegraphics[width=\textwidth]{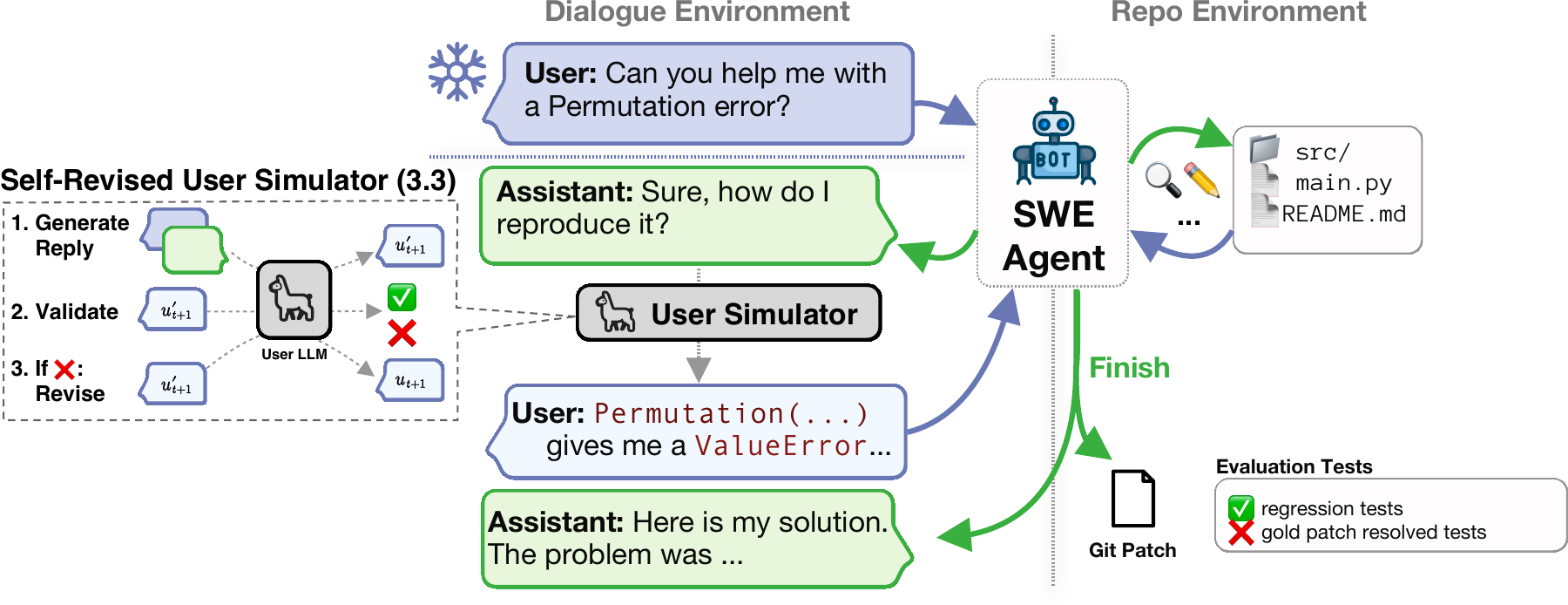}
    \caption{Overview of our benchmark setup and user simulator. The user never interacts with the code, and performs software engineering tasks strictly through dialogue. For each SWE problem, the dialogue starts with an initial fixed query, and proceeds with an online evaluation of agents with a user simulator which replies to each agent message. To ensure faithful simulation of the user, we include a self-revision step for the user (\S\ref{sec:user-simulation}), where a generated reply $u'_{t+1}$ is validated and then revised if necessary.}
    \label{fig:main-problem}
\end{figure*}

While real-world coding agent use is often dialogue-driven, to our knowledge, no benchmark exists for systematically evaluating the dialogue capabilities of coding agents at repository scale. 
Previous works have evaluated dialogue systems for producing single functions \cite{chaurasia_dialog_2017}, but are not suitable for repository-level tasks.
Recent work evaluates the ability of agents to resolve ambiguous problem specifications \cite{vijayvargiya2026ambigswe}, but does not fully address dialogue-driven use of coding agents.

In this paper, we introduce \textbf{\datasetname}: a benchmark dataset for evaluating the ability of coding agents to resolve real-world software engineering problems through dialogue with a user.
We evaluate state-of-the-art coding agents and find that better coding models do not always correspond better interactive agents in the dialogue setting.
Our contributions are as follows:
\squishlist
    \item We introduce a \textit{dialogue-driven} benchmark for coding agents in which they must solve real-world SWE tasks via dialogue with a user, rather than being given a complete problem specification upon task start (\autoref{sec:benchmark}).
    \item To do this, we design a novel, persona-grounded user simulator for online evaluation of coding agents through dialogue (\autoref{sec:user-simulation}).
    \item We evaluate state-of-the-art coding models and agents in this new dialogue setting, and find that stronger coding models are not always stronger dialogue systems (\autoref{sec:experiments}-\autoref{sec:results}).
    \item We support our task evaluation with automatic evaluation of dialogue quality using a novel LLM-as-a-Judge (\autoref{sec:nat-and-coherence}).
    \item We propose a new, dialogue-theory informed coding agent with the best average performance on this task (\autoref{sec:methods}).
\squishend

\section{Related Work}
\label{sec:related-work}
The rise of language agents \cite{su-etal-2024-language} -- AI agents whose policies are mediated by LLM reasoning -- has enabled a shift in text-to-code research from writing simple functions \cite{chen_evaluating_2021} to real-world software engineering (SWE) \cite{jimenez_swe-bench_2024, aleithan_swe-bench_2024}.
While there has been substantial progress made on SWE tasks through advances in language models \cite{rastogi2025devstralfinetuninglanguagemodels, qwen3technicalreport}, agent design \cite{wang_openhands_2024, yang_swe-agent_2024}, and fine-tuning techniques \cite{pan_training_2024, wei_swe-rl_2025, yang2025swesmith}, much of this work considers only the fully automated SWE setting, which burdens the user with providing a complete and correct specification and ignores the potential of human-agent interaction.
The research community has begun to argue for the need to model such human-agent interactions \cite{wang_position_2026}, but evaluating such interactive settings remains an open challenge.

Two recent works evaluate human-agent interaction in the SWE setting.
\citet{vijayvargiya2026ambigswe} the ability of coding agent in resolving ambiguity from incomplete problem specification through clarifying questions. 
While interactive, this setup still presumes a specification document as input, with interaction used only to resolve ambiguity. 
Our evaluation differs in that our task begins with user utterance and unfolds entirely through dialogue. We compare our proposed agent to theirs in our evaluation.
\citet{zhou2026tomsweusermentalmodeling} design an evaluation for multi-session coding interactions with a coding agent, evaluating the ability of agents to honoring user-level preferences across sessions. Our evaluation differs in that our setup is dialogue-driven with a single SWE task, and our user simulator does not interact with code and is not given knowledge that could contain the task solution, such as issue comments \cite{aleithan_swe-bench_2024}.

Prior to the advent of coding agents, other works have considered dialogue for \textit{function-level} text-to-code problems.
Some works use the gold solution to a text-to-code problem to create a dialogue for that problem, by crafting clarification question \& answer pairs \cite{li_python_2023}, simulating a user's feedback \cite{pan_when_2025, wang_mint_2023}, or addressing uncertainty in a model's predictions \cite{chaurasia_dialog_2017}.
\citet{rakotonirina_tools_2025} evaluate LLMs in multi-session coding interactions across several \textit{function-level} coding tasks.
Our work instead addresses dialogue for \textit{repository-level} SWE tasks, and importantly, never conditions the simulated user on gold knowledge, like solutions or test cases.

\section{\datasetname}
\label{sec:benchmark}

In \autoref{sec:preliminaries}, we review coding agents and the fully-autonomous evaluation setup, as introduced in SWE-Bench \cite{jimenez_swe-bench_2024}. 
We then describe how we re-formulate the fully-autonomous evaluation into an interactive setup which \textit{requires dialogue} (\autoref{sec:problem-design}) 
and provide further detail on the user simulator we design to support our dialogue-driven evaluation (\autoref{sec:user-simulation}) and our evaluation metric (\autoref{sec:eval-design-task-resolution}).

\subsection{Preliminaries}
\label{sec:preliminaries}

In the fully-autonomous SWE evaluation, a system generates a repository-level code change (e.g. git patch) from a problem description and initial repository. This patch is then evaluated using unit tests. 
The dominant approach to this task uses a coding agent or `SWE Agent' \cite{yang_swe-agent_2024}, an LLM-powered agent which operates directly within a programming environment to iteratively explore, edit, and execute code.
Given an initial repository $\repo$ and problem specification $\issuetext$, A coding agent is initialized in an observation-action loop, beginning with a first observation $o_1$ containing the problem specification $\issuetext$.
At each time step $t$, the agent executes an action $a_t \in \mathcal{A}$, such as editing a file or running a test, and receives a corresponding observation $o_{t+1} \in \mathcal{O}$, such as execution results, where $\mathcal{O}$ and $\mathcal{A}$ represent the agent's observation space and action space, respectively.
A special action `finish' terminates the loop and prepares a patch $\patch$ from the edited files for evaluation.
Each problem is evaluated by applying the generated patch $\patch$ to the base repository $\repo$ and then executing tests associated with the task.
A task is considered resolved if all tests pass after the patch is applied, and the evaluation metric for the benchmark is the percentage of resolved tasks.

\subsection{Dialogue Problem Design}
\label{sec:problem-design}

We propose a benchmark in which coding agents must resolve real-world software engineering problems through goal-oriented dialogue with a user.
We re-formulate SWE-Bench Verified \cite{chowdhury_introducing_2024} into this benchmark as follows (overview in \autoref{fig:main-problem}).
First, we replace the full issue text $\issuetext$ in the first observation $o_1$ with an initial user query $u_1$.
Next, we situate agents in an environment with an action space which supports both programming and dialogue with a simulated user.
Specifically, we augment the action space $\mathcal{A}$ of each agent with an additional action `\texttt{message\_user}', which takes a single argument containing the message to communicate with the user, and yields an observation $o_{t+1}$ containing their response.
We simulate each user using an LLM, with knowledge grounded in the full issue text $\issuetext$, described further in \autoref{sec:user-simulation}.
As before, the agent's observation-action loop terminates with the `finish' action, yielding a patch $\patch$, which can be evaluated as resolving or not resolving the user's issue using execution tests.

\subsection{Simulating Users}
\label{sec:user-simulation}
To reduce variance across runs in our benchmark, we design a fixed initial user query $u_1$ for each problem.
Following this, we design a simulator for producing faithful and realistic user replies to follow-up queries from the agent.  To enable stable future evaluation on the benchmark, the user simulator is open-source and released as part of the benchmark.

\paragraphsquish{Crafting initial queries}
We design each initial query $u_1$ to faithfully reflect the intent given in the problem specification but omit key details needed for resolution, using the following semi-automated approach.
First, we prompt an LLM with the Github issue title and brief instructions for paraphrasing it into an initial dialogue query.
Following this, we use a revision prompt to remove any critical details from the query that an agent might be able to exploit to avoid engaging in dialogue altogether.
Finally, we manually review each query to verify it is faithful to the intent described in the problem specification.
\autoref{app:user-sim-details} provides further details on our prompting approach and manual review.
While our manual review ensures a high-quality evaluation set, we find only 13\% of generated queries require any modification, suggesting our pipeline could be fully automated in training or development settings.

\paragraphsquish{Simulating follow-up replies with Self-Revision}
For any timestep $t\ge1$, we prompt an open-weights LLM  $\userllm$ to simulate the user's reply to an agent's message $u_{t+1}$.\footnote{We use a quantized version of LLaMa 3.3 70B \cite{grattafiori_llama_2024} available at \href{https://huggingface.co/shuyuej/Llama-3.3-70B-Instruct-GPTQ}{https://huggingface.co/shuyuej/Llama-3.3-70B-Instruct-GPTQ}}
Specifically, we define a system prompt for the user simulator conditioned on the user's knowledge for that problem, given in the complete issue text $\issuetext$, as well as a persona for grounding behavior, detailed below.
Following this, all dialogue messages for that problem are included in the context for generating a candidate for the next user utterance $u'_{t+1}$.
To ensure our user simulator behaves realistically for the setting, we use a revision step, in which a candidate user utterance $u'_{t+1}$ is validated by the same user LLM $\userllm$, checking for potential hallucinations or violations of the task instructions.
For example, since our simulated user is unable to run new code on its own, we use the revision step to identify any replies in which the user claims to do so. 
We also test for directly detectable violations, such as exceeding the length limit in the reply.
If any errors are found, we use a final prompt to revise the candidate utterance before it is returned to the agent as an observation.
See \autoref{app:user-sim-details} for further details.

\paragraph{User Personas}
To improve diversity of dialogues simulated in our benchmark we assign each problem a user persona, following \citet{rakotonirina_tools_2025}.
For each problem, we sample a name and hand-crafted persona description which will be used to influence responses given by our user simulator.
Names and persona descriptions are given in \autoref{tab:persona-names} and \autoref{tab:persona-descriptions} of \autoref{app:user-sim-details}, respectively.

\input{latex/table_1} %

\subsection{Evaluation Metric}
\label{sec:eval-design-task-resolution}
The principal metric for our benchmark is the resolution rate, or the percentage of  dialogues which result in a correctly resolved SWE task.
To evaluate this, we use the execution tests corresponding to each problem from SWE-Bench Verified \cite{chowdhury_introducing_2024}.
A dialogue is considered successful if the conversation results in a submitted patch $p$ which passes all associated execution tests.

\section{Schema Guided Coding Agents}
\label{sec:methods}

To address the unique demands of our dialogue benchmark, we propose a novel schema-guided coding agent which adaptively plans dialogue moves using \textit{dialogue schemas}, or ``structured representations of the prototypical sequence of events in a dialogue'' \cite{kane_schema-guided_2022}.
Specifically, we use prompts to instruct our agent to build and maintain its own structured representation of the dialogue state in order to guide its questions, code exploration, and patch generation.

\begin{figure}
    \centering
    \includegraphics[width=\columnwidth]{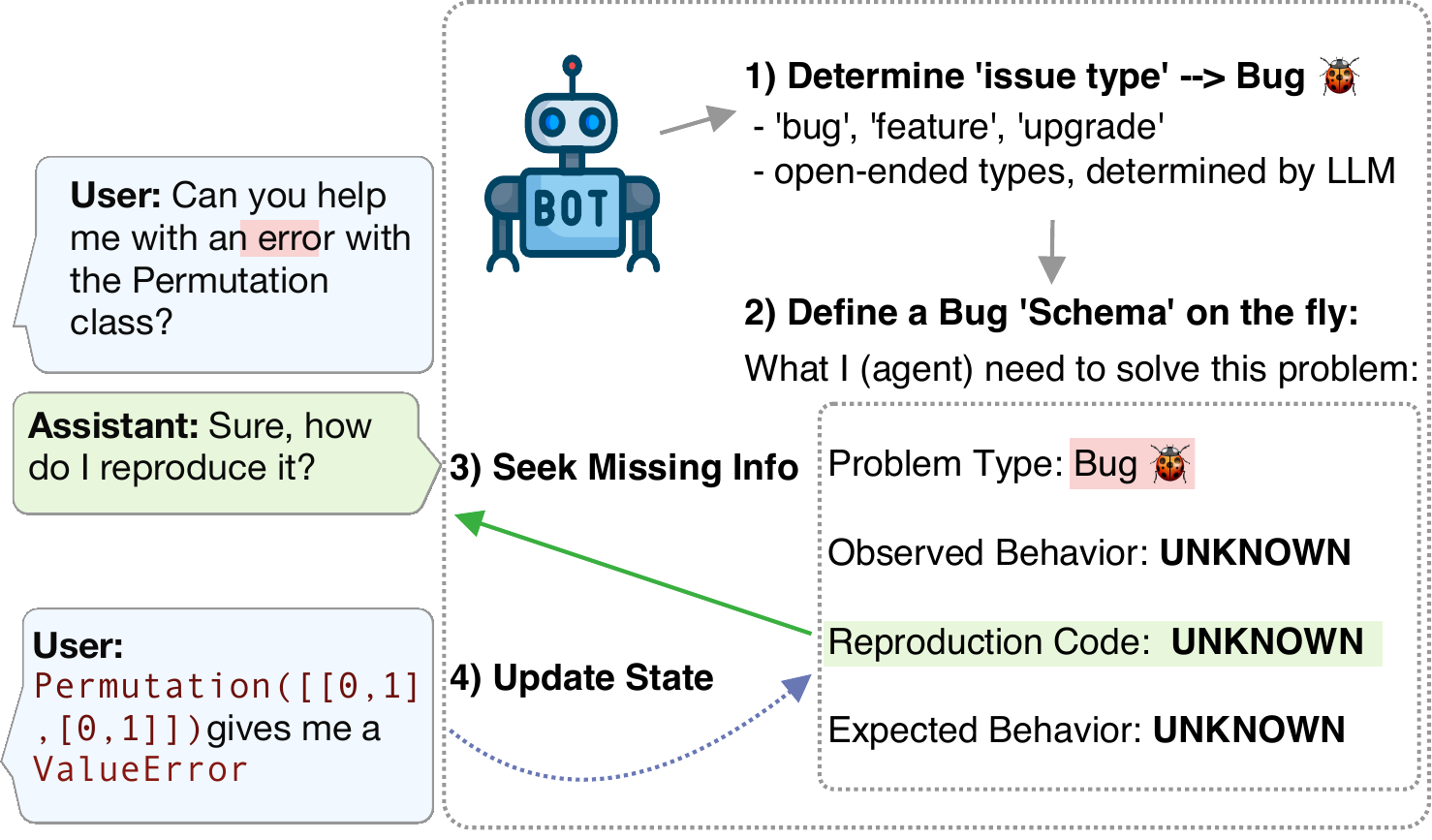}
    \caption{Workflow for our schema-guided SWE Agent}
    \label{fig:schema-guided-agent}
\end{figure}

Using prompts, we instruct our agent to conduct a schema-guided dialogue as follows (overview in \autoref{fig:schema-guided-agent}).
First, we instruct the agent to determine the type of issue the user presents and draft a schema with keys and values for the critical details needed to resolve the issue. 
The possible issue types, keys, and values in this schema are all determined by the agent.
For example, if the user seems to be describing a bug, the agent might draft a schema with an issue type of `bug', with keys for `actual behavior', `expected behavior', and `reproduction steps.' 
We instruct the agent to mark values which have not yet been discussed as UNKNOWN, and fill these in using dialogue, until enough details are gathered to begin solving the problem. 
The agent is instructed to maintain this dialogue state as it explores code, makes changes, and verifies a solution.
We implement our agent using the OpenHands Agent SDK \cite{wang2025openhandssoftwareagentsdk}, using the default tools for coding agents, including file editing, bash tools, and finishing.

\section{Experimental Setup}
\label{sec:experiments}

In this section, we describe our evaluation of closed and open-weight LLMs on our benchmark.
For all experiments, we evaluate systems with our dialogue transformation of the 500 problems from SWE-Bench Verified \cite{chowdhury_introducing_2024}.

\paragraphsquish{Selected Models}
We evaluate a representative set of closed and open-weight models based on their performance on SWE-Bench.
For proprietary models, we evaluate GPT-5 and GPT-5-mini \cite{openai_gpt-5_2025}.
We also evaluate the open-weight models Qwen3 Coder 30-A-3B \cite{qwen3technicalreport} and Devstral 2 Small \cite{mistral_introducing_2025}, both trained for agentic coding.

\paragraphsquish{Benchmark Setup} For all experiments, we instantiate our user simulator with a quantized version of LLaMa 3.3 70B.\footnote{\href{https://huggingface.co/shuyuej/Llama-3.3-70B-Instruct-GPTQ}{https://huggingface.co/shuyuej/Llama-3.3-70B-Instruct-GPTQ}}
We also limit each agent to 100 steps per problem instance.
For each agent, we include an additional \texttt{message\_user} tool, providing an additional interface for communicating with the user.\footnote{This tool takes a single argument \texttt{<message>} and returns the user's response. An assistant message without a tool call is similarly treated as a dialogue message.}

\paragraphsquish{Baselines and Agents} For each model, we consider the following baseline agents: (1) OpenHands \cite{wang_openhands_2024}, as an off-the-shelf coding agent, and (2) OH Interactive \cite{vijayvargiya2026ambigswe}, as a baseline designed to interact with a user to resolve ambiguity.
We also evaluate each model with our schema guided agent (\autoref{sec:methods}). Importantly, all agents share the same set of available tools.

\section{Results}
\label{sec:results}

We present the performance of each approach on our benchmark in \autoref{tab:main-results}.
In addition to our core task metric (\% Resolved), we report the average number of dialogue turns, agent steps, and cost in dollars per problem session.\footnote{See \autoref{app:cost-calculation} for details on cost calculations.}
We find that our schema-guided agent achieves the highest average resolve rate across all models (46.9\%), outperforming OpenHands (32.9\%) and OH Interactive (44.1\%), at the lowest average cost.
While our schema-guided agent typically makes more use of dialogue, this is not accompanied with a corresponding increase in the number of total steps.
Surprisingly, we find that performance of GPT-5 mini rivals that of GPT-5 at a fraction of the cost, and that strong coding ability does not always correspond with dialogue ability.
We investigate this further in a few ways throughout the paper.
First, in \autoref{app:stratify-by-diff}, we stratify performance by engineering difficulty, finding that while the larger GPT-5 performs best on harder engineering problems, it under-performs on simpler tasks.
We find this is in some cases due to dialogue failures, such as asking too many unnecessary questions or failing to follow up on a missing detail, and present a case study as an example in \autoref{sec:case-study}.
Finally, we note that relative to GPT-5, GPT-5 mini dialogues rate as more natural and coherent (\autoref{sec:nat-and-coherence}).
Altogether, these results demonstrates that \textit{\textbf{strong coding ability does not always coincide with dialogue ability}}, highlighting the importance of a dialogue benchmark for coding agents.

\paragraphsquish{Information-Seeking Drives Task Resolution}
In \autoref{fig:info-seek-moves}, we look more closely at the number of information seeking dialogue moves used by each agent and its relationship with Resolve Rate.
Using an LLM classifier, we determine whether a given dialogue message from agent to user seeks information about the problem, rather than serving only other rhetorical functions such as summarizing changes made, offering further assistance, or providing a greeting/conventional closing.
We use GPT-5-mini to classify each agent message as containing a question-info-request dialogue act \cite{jurafsky_switchboard_1997} or not, and find strong agreement with human annotation (Cohen's $\kappa$ = 0.87).
We note a few interesting findings.
First, the best performing agent for a given model consistently uses more information-seeking dialogue moves.
Second, the off-the-shelf agent framework OpenHands rarely uses information seeking moves, suggesting that intervention to the agent or model is necessary to support multi-turn negotiations of a user's intent in the dialogue setting. 
Finally, our schema-guided agent typically uses the highest number of information seeking moves, with the exception of Devstral 2 Small.

\input{latex/table.resolved_vs_avg_info_seek}

\section{Analyzing Dialogue Quality}
\label{sec:nat-and-coherence}

To complement our benchmark evaluation, we propose automatic metrics of dialogue quality for the coding agent. 
Following \citet{kazi_large_2024}, we evaluate our agents along two dimensions, Naturalness and Coherence.
For each measure, we devise an LLM-as-a-judge which assesses the performance of an agent at the dialogue level, and validate it's judgments with human annotation. 
For both metrics, we implement our LLM-as-a-Judge using Gemma 4 31B-IT \cite{gemma_4_model_card}.

\subsection{Naturalness} 
We analyze naturalness as the degree to which an agent is easy to understand and converse with. To assess naturalness, we use the following three point scale:
\begin{enumerate}
    \item \textbf{Low naturalness (1)}: this agent does not communicate in an understandable way, or has clear communication issues that a user may find frustrating.
\item \textbf{Medium naturalness (2)}: this agent communicates mostly in an understandable way, but there are some minor issues that make the dialogue feel slightly less natural.
\item \textbf{High naturalness (3)}: this agent communicates clearly and concisely, and in way user's would find easy to communicate with and understand.
\end{enumerate}

\begin{figure}
    \centering
    \begin{subfigure}{\columnwidth}
        \centering
        \includegraphics[width=.82\columnwidth]{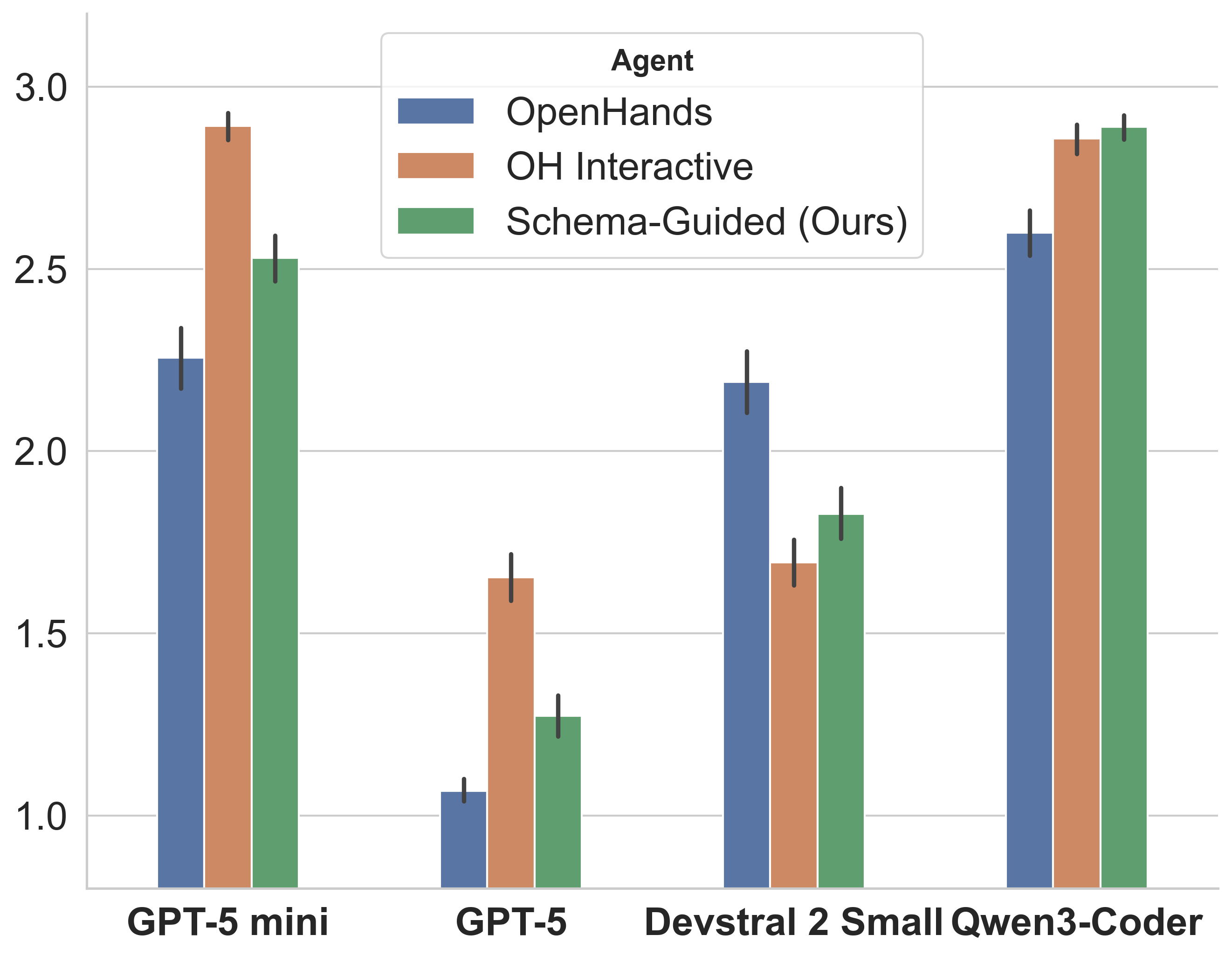}
        \caption{Dialogue naturalness}
        \label{fig:nat-full-llm-results}
    \end{subfigure}
    \begin{subfigure}{\columnwidth}
        \centering
        \includegraphics[width=.82\columnwidth]{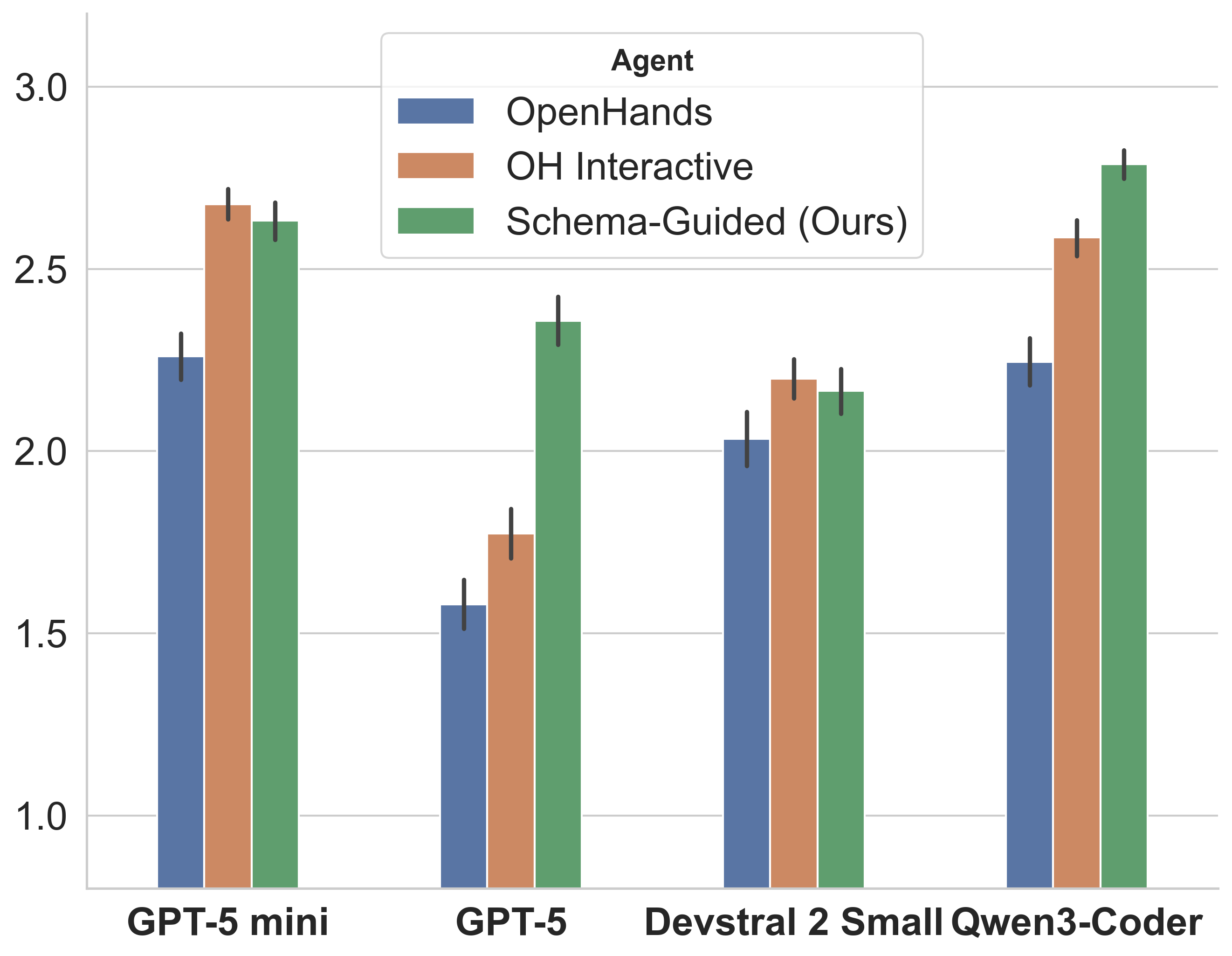}
        \caption{Dialogue coherence}
        \label{fig:coh-full-llm-results}
    \end{subfigure}
    \caption{LLM-as-a-judge ratings for naturalness and coherence (1-3) with 95\% confidence intervals. }
    \label{fig:nat-coh-full-llm-results}
\end{figure}

\paragraphsquish{Results} We present results from our LLM-as-a-Judge in \autoref{fig:nat-full-llm-results}. Across systems, we find more variance in naturalness from choice of model than choice of agent. Notably, GPT-5 suffers from low naturalness, as it often fails to close a dialogue correctly or repeats internal portions of its system prompt to the user. We explore this further as a case study in \autoref{sec:case-study}.
Alternatively, Devstral's low naturalness is due to failure to participate in dialogue: in 23-34\% of dialogues, devstral never responds to the user, exiting on max calls.
Finally, we observe no clear relationship between naturalness and resolution rate, suggesting the usability of a coding agent in dialogue cannot be measured by task success alone.

\subsection{Coherence} 
We analyze coherence as the degree to which an agent's dialogue moves guide the dialogue on a logical path toward solving the user's task, on a 1-3 scale. Following \citet{kazi_large_2024}, we derive the agent's score from two criteria:
\begin{enumerate}
    \item \textbf{Local coherence}: whether each agent turn logically follows from the preceding messages (i.e. is on task and not redundant).
\item \textbf{Global coherence}: whether the agent's contributions collectively steer the conversation solving the user's task, including seeking clarification or detail as needed and closing the dialogue appropriately.
\end{enumerate}

\noindent If neither criteria are met, the agent scores 1. If only one of the two criteria are met, the agent scores 2. An agent which meets both criteria scores 3.

\paragraph{Results} We present results from our LLM-as-a-Judge in \autoref{fig:coh-full-llm-results}. We observe stronger differentiation between agents on coherence than naturalness, suggesting that agent design can more strongly influence a system's ability to construct a logical dialogue path toward issue resolution. 
Our schema-guided agent rates the most coherent for two out of four models with statistical significance.\footnote{Determined with a wilcoxon signed rank test, $p<10^{-5}$} For GPT-5 mini and Devstral, the difference between the leading OH Interactive and our schema-guided agent is not statistically significant.

\subsection{Evaluating the LLM-as-a-Judge}

To evaluate our LLM-as-a-judge we conduct a human annotation of 30 dialogues per model-agent pair for a total of 360 dialogues. 
We score each system using the same 3-point scale for naturalness and coherence.

We evaluate our judge with two measures of agreement.
First, we quantify the agreement between our judge and human ratings using a quadratically-weighted Cohen's $\kappa$ \cite{cohen_weighted_1968}.
Second, we evaluate our judge's ability to correctly rank competing systems, following \citet{kocmi-etal-2021-ship}. We begin by identifying pairs of evaluated systems whose human ratings differ significantly under a Wilcoxon signed rank test \cite{pratt_remarks_1959}. We then define ranking accuracy as the percentage of these pairs for which the judge's ranking agrees with the human ranking.

\begin{table}[]
    \centering
    \resizebox{0.9\columnwidth}{!}{
    \begin{tabular}{lrr}
    \textbf{LLM-Judge} & \textbf{Cohen's $\kappa$} & \textbf{Rank Accuracy ($n$)} \\
    \hline
        Naturalness & 0.70 & 100\% \small{($n$=42)} \\
        Coherence &  0.51 & 84.3\% \small{($n$=51)} \\
    \hline
    \end{tabular}
    }
    \caption{Agreement of our LLM-as-a-judge with human ratings, measured using quadratic weighted Cohen's $\kappa$ and ranking accuracy over the $n$ system pairs with a statistically significant difference in human ratings}
    \label{tab:human-llm-agreement}
\end{table}

We report the agreement measures between our LLM-as-a-judge and human ratings in \autoref{tab:human-llm-agreement}. \autoref{fig:nat-human-results} and \autoref{fig:coh-human-results} of \autoref{app:nat-coh-annotation} provide our human ratings for naturalness and coherence, respectively.
For naturalness, we find substantial agreement between our LLM-as-a-judge and human ratings ($\kappa$=0.70). Our naturalness judge also ranks systems by naturalness with 100\% accuracy.
For coherence, we find moderate agreement ($\kappa$=0.51), due in part to the complex judgment required to assess an entire dialogue's logical flow.
Despite only moderate agreement, our coherence judge still achieves a rank accuracy of 84.3\%.

\section{Ablations \& Analyses}
\subsection{Ablating our User Simulator}
\label{sec:user-sim-ablation}
We verify that our benchmark evaluates multi-turn problem solving through dialogue using an ablation that removes the user simulator after the first turn. 
As before, we seed the dialogue with the user's fixed initial query, but then return ``the user is not available.'' for any subsequent user messages.
The agent must therefore solve the task without any interaction.
For this ablation, we sample 50 problems from the evaluation set stratified by difficulty as annotated by \citet{chowdhury_introducing_2024}. We use our schema-guided agent (\autoref{sec:methods}) with each model.
\autoref{tab:user-sim-ablation} shows that removing the user simulator reduces performance significantly across models, indicating that success on our benchmark requires multi-turn dialogue capabilities.

\input{latex/table.user_sim_ablation}

\subsection{Evaluating our User Simulator}
\label{sec:analysis-validating-user-sim}

We evaluate whether our user simulator can effectively serve as a proxy for real-world users through human annotation.
To do this, we evaluate our user's goal-oriented behavior along three binary dimensions: 
\begin{enumerate}
    \item \textbf{Faithfulness}: is the user faithful to the intent and knowledge in the issue specification?
    \item \textbf{Goal Adherence}: does the user reliably pursue their stated goal, and not get side-tracked?
    \item \textbf{Environment Adherence}: does the user respect the limitations of their environment, and not hallucinate abilities like running code?
\end{enumerate}
\noindent Each criterion is annotated at the utterance level as met or not met using the same annotation guidelines.
We evaluate both our final user simulator, and an ablation which removes our revision step described in \autoref{sec:user-simulation}.
We sample 120 dialogues across model and agent choices, and report results in \autoref{tab:user-sim-eval-goal-oriented}.
We find that our best user simulator scores highly across all three dimensions at both the turn and dialogue level, demonstrating strong goal-oriented behavior.
Without our revision step, the user simulator performs significantly worse, particularly with environment adherence. 
Further, our full user simulator scores perfectly on all dimensions in 97.5\% of dialogues, where our simulator without revision only scores perfectly on 82.5\%.

\begin{table}[h]
    \centering
    \resizebox{\columnwidth}{!}{
    \begin{tabular}{lrrrr}
        \toprule
        & Faith. ($\uparrow$) & Goal Adh. ($\uparrow$) & Env. ($\uparrow$) & \textbf{All ($\uparrow)$} \\
        \midrule
        \multicolumn{5}{l}{\textit{\textbf{Turn-Level}}} \\
        User Simulator (\autoref{sec:user-simulation})      & \textbf{99.7} & \textbf{99.7} & \textbf{99.7} & \textbf{99.1} \\
        \quad $-$ self-revision    & 97.8 & 89.7 & 92.2 & 84.1 \\
        \multicolumn{5}{l}{\textit{\textbf{Dialogue-Level}}} \\
        User Simulator (\autoref{sec:user-simulation})      & \textbf{99.2} & \textbf{99.2} & \textbf{99.2} &  \textbf{97.5} \\
        \quad $-$ self-revision    & 95.0 & 92.5 & 87.5 & 82.5 \\
        \bottomrule
    \end{tabular}
    }
    \caption{Human evaluation of faithfulness, goal adherence, and environment adherence in our User Simulator (\autoref{sec:user-simulation}). Scores are percentage of utterances (turn-level) and dialogues (dialogue-level) where each criteria is met. All ($\uparrow$) is the percentage of turns or dialogues where all criteria are met. Our user meets each criteria in over 99\% of dialogues, with 97.5\% of dialogues scoring perfectly. Without self-revision (\autoref{sec:user-simulation}), only 82.5\% of dialogues are defect free.}
    \label{tab:user-sim-eval-goal-oriented}
\end{table}

\section{Case Studies}
\label{sec:case-study}

We analyze system behavior in our benchmark through case studies.

\paragraph{Schemas Guide Reasoning} First, we note an example in which our schema-guided agent outperforms our strongest baseline (OH Interactive) because it is able to surface a key details about the expected behavior through dialogue. 
\autoref{fig:case-study-ours-vs-baseline} in \autoref{app:case-study-details} shows this example in detail.
In this example, the user intends to modify a function to accept an argument as either (1) None, mapping it to an empty \texttt{set}, or (2) as-is, asserting that the passed argument is of type \texttt{set}. 
While both agents ask about what types to accept, the schema-guided agent reasons about the expected behavior in it's schema explicitly, and further considers the expected behavior when the caller's argument is not a \texttt{set}.
The schema-guided agent proposes one solution (coercing any iterable to a \texttt{set}), which prompts the user to respond with the correct alternative solution: raising an assertion error if the argument is not a \texttt{set}.
The baseline instead assumes it is the \textit{caller's} responsibility to ensure the argument is a \texttt{set}, violating the user's true intent.

\paragraph{Verbose Models Burden Users} We also notice that while thorough, some models commonly burdens the user with unnecessary questions.
In particular, we notice this across agents with GPT-5, in part leading to the longer dialogues we observe in \autoref{tab:main-results}. 
\autoref{fig:case-study-gpt-5-vs-mini} in \autoref{app:case-study-details} provides an example dialogue comparing GPT-5 and GPT-5-mini, both using our schema-guided agent. 
The user's problem is a `$<15$ min fix' and has a very simple solution: recognizing improper inputs and throwing an exception. 
Where GPT-5-mini asks two clear and useful questions, GPT-5 asks several more, many of which are unnecessary. 
In response to a long list of questions from GPT-5, the user selectively answers only two.
GPT-5 declines to follow-up on an unanswered question about expected behavior for improper inputs, and instead implements a solution which ignores such inputs rather than raising an exception.

\section{Conclusion}
\label{sec:conclusion}

We introduce Dialogue-SWEBench, an evaluation of coding agents on real-world software engineering tasks through multi-turn dialogue with a simulated user.
We hope our work inspires future work on the intersection of coding and dialogue capabilities needed for real-world interactive coding agent use.

\section{Limitations}
\label{sec:limitations}

We build our benchmark using real-world problems from SWE-Bench Verified \cite{chowdhury_introducing_2024}. 
The annotations provided in the the verified set allow us to be confident that each task's problem specification is complete and correct, and thus solvable in principle. 
However, this design choice means our benchmark inherits the distribution of SWE-Bench Verified, which skews toward Python repositories and particular issue types that might not cover the full distribution of real-world coding agent uses.
Future work will be needed to assess coding agent dialogue capability in a greater variety of repository-level task domains.

\section*{Acknowledgments} 
The authors were supported in part by the NSF National
AI Institute for Student-AI Teaming (iSAT) under
grant DRL 2019805. The opinions expressed are
those of the authors and do not represent views
of the NSF. 
We are thankful for the computing resources provided by the Pacific Research Platform's Nautilus cluster, supported by the National Science Foundation under Award Numbers CNS-1730158, ACI-1540112, ACI1541349, OAC-1826967, the University of California Office of the President, and the University of California San Diego’s California Institute for Telecommunications and Information Technology/Qualcomm Institute.

\bibliography{custom}

\appendix

\section{User Simulation Details}
\label{app:user-sim-details}

Here we provide further details on the design of our user simulator.
For both generating initial queries and simulating follow up responses, we use a quantized version of LLaMa 3.3 70B \cite{grattafiori_llama_2024}.\footnote{\href{https://huggingface.co/shuyuej/Llama-3.3-70B-Instruct-GPTQ}{https://huggingface.co/shuyuej/Llama-3.3-70B-Instruct-GPTQ}}

\paragraph{Generating Initial Queries}
Our semi-automated procedure for producing initial queries has three key steps.
First, we prompt the LLM $\userllm$ with the title of the Github issue to produce a candidate initial query for the dialogue, $u'_1$.
This prompt is in \autoref{fig:user-sim-prompt-draft-query}.
Following this, we prompt the LLM to revise the candidate $u'_1$ to ensure it is faithful to the user's original intent, sufficiently incomplete to require dialogue, and meets the requirements of our evaluation setting.
This revision prompt is in \autoref{fig:user-sim-prompt-revise-query}.
Finally, we manually review the initial queries to ensure they are clear and faithful to the user's original intent.
In preparing our evaluation set, 13\% of the problems required minor manual revision, to ensure they had an issue resolution based framing, rather than asking an open ended question.

\paragraph{Generating Follow-Up Responses}
To produce a follow-up query $u_{t+1}$, we first prompt the user LLM $\userllm$ to produce a candidate reply $\hat{u}_{t+1}$. We combine the system prompt detailed in \autoref{fig:systemprompt} with the dialogue history thus far, with \textit{roles reversed}.
This way, the LLM $\userllm$ is tasked with producing a message with the  `assistant' role, matching its instruction tuning setup, before being returned to our agent as a user message.
Following candidate generation, we use a \textbf{self-revision step}, in which we prompt the user LLM with the candidate $\hat{u}_{t+1}$ to check for possible hallucinations or violations of our setting. \autoref{fig:revision-prompt} details the prompt for detecting and classifying such violations.
Finally, if any violations are recorded, the LLM $\userllm$ is prompted to revise $\hat{u}_{t+1}$ before it is returned to the agent as an observation $o_{t+1}$.

\paragraph{Sampling Personas}
Inspired by the approach in \citet{rakotonirina_tools_2025}, we aim to evaluate agents against a more diverse and varied set of users using personas.
We define five persona descriptions for users in our benchmark, given in \autoref{tab:persona-descriptions}.
For each problem in our evaluation set, we sample a description and instantiate it with a name sampled from \autoref{tab:persona-names}, using the same names as in \citet{rakotonirina_tools_2025}.
The resulting persona text is then used to influence our user simulators behavior in the system prompt.

\begin{figure}[htbp]
\centering
\begin{lstlisting}[style=columnpromptstyle]
This is the title of an issue on Github that a user needs help with. Can you re-phrase the title as a question or other natural beginning of a dialogue it will have with a pair programmer? Be brief and capture the essence of the issue, the pair programmer will ask questions to get details and resolve ambiguity.
Where possible, avoid "why" or "let's discuss" questions.

Here is a description of the user you are trying to emulate:

~{{ persona }}~

Here is the title of the issue:

Issue Title: ~{{ issue_title }}~

Please phrase as a natural query, for example: "Can you help me figure out why `iam.physical` is returning NaN for an AoI greater than 90?"
\end{lstlisting}
\caption{Prompt used for producing a draft initial query $u'_1$ for initializing our user simulator.}
\label{fig:user-sim-prompt-draft-query}
\end{figure}
\begin{figure}[htbp]
\centering
\begin{lstlisting}[style=columnpromptstyle]
Now here is the full issue context. The initial query will become the user's initial query in a simulated benchmark for software engineering dialogues.

Please check carefully for two possible issues:

- Misconceptions: does the query mis-represent the intent of the user? Intents should be broad: 'fix a bug', 'safely upgrade a dependency', etc. Missing details are not misconceptions: details should be surfaced through the multi-turn dialogue.
- Over-detailed intents: does the query too much information, that should instead come from dialogue with the user? If a query can be re-stated in a way that removes details that are important to solving the problem, without leading to a change in the user's broad intent (fixing a bug, adding a feature, etc.), then it should be simplified, with details to follow in dialogue.

Formatting: You should verbally think through and decide whether to make any changes first. After this, end your response with "Query: <query>"

Here is the full problem statement. We want as few details as possible from this in the initial query without destroying its meaning.

~{{ problem_statement }}~

Query: ~{{ initial_query }}~
\end{lstlisting}
\caption{Prompt used for automatic revision of a draft initial query $u'_1$ for initializing our user simulator.}
\label{fig:user-sim-prompt-revise-query}
\end{figure}

\begin{figure*}[htbp]
\centering
\begin{lstlisting}[style=promptstyle]
SETTING:

Your task is to simulate a Github User engaging in a dialogue with an autonomous software engineer to solve a problem.
You must naturally, accurately, and authentically behave as a user seeking a solution to a software engineering problem.
Your task is to pose the problem to an engineer in a natural dialogue, and clearly and concisely answer any questions from the engineer.

To do this, you need to always adhere to the following rules and boundaries concerining your persona, environment, and knowledge:

ABOUT YOUR PERSONA:

~{{ persona }}~

RULES FOR PERSONA:
1. Success at this task REQUIRES that you behave according to this PERSONA, even when you would typically behave differently!
2. NEVER describe yourself as a simulator, break character as a user, or acknowledge that you have a problem statement you are consulting. Your task is to AUTHENTICALLY behave as a user that wishes to solve the problem in real-time, without a pre-written problem statement.
3. ALWAYS be as brief and natural as possible: do not overwhelm the engineer with too much information! Remember that real-world users are not wordy, they aim to type as little as possible.
4. If the engineer has overwhelmed you with many questions in one message, you do not need to answer all of them. They will ask again if needed.

ABOUT YOUR ENVIRONMENT:

You are engaging with the autonomous engineer only through dialogue (like a slack DM). Follow these rules and boundaries with respect to your environment:

The engineer is a maintainer of the ~{{ repo }}~ repository.
Your problem is described below and must be solved by the engineer through some change to the ~{{ repo }}~ repository.

RULES FOR YOUR ENVIRONMENT:
1. Always remember that you are situated only in dialogue with the software engineer: you do not have a terminal, IDE, access to the repository (neither ~{{ repo }}~ nor the repo where the problem surfaced), or the ability to complete any task or follow-up task on your own. You may ask the engineer to run code in their environment if necessary. Do NOT pretend to have these or make promises about follow-up tasks you can complete.
2. Use only your provided knowledge to participate in the dialogue. While you should never break character, you can be forthright about the limitations of your environment ("e.g. I can't run any code right now", "I can't see that right now", "I don't have that information right now")
3. You will be given the full text of a Github Issue below. NEVER share it directly with the engineer, but instead use it as a knowledge source for authentically behaving as the user.
4. The engineer is tasked with all implementation and any testing, if necessary, communicating with you only through dialogue. Their task is not complete until the problem has been SOLVED in the ~{{ repo }}~ repository.
5. The engineer can only provide you information through dialogue. If they refer code or something else you cannot see, ask them to message it to you.
6. You may ask the engineer to run code in the ~{{ repo }}~ repository if necessary, but you may not run code yourself. You may refer to code that was run previously and described in complete detail in the problem statement, but you can

YOUR KNOWLEDGE FOR THIS PROBLEM (Full Github Issue/Problem Statement):

~{{ full_problem_statement }}~

RULES FOR KNOWLEDGE AND DIALOGUE:
  1. ALWAYS be clear with your answers using excerpts or paraphrases of the full problem statement given above.
  2. NEVER make up an answer to a question you cannot answer by consulting the problem statement. You can respond to these questions with "I'm not sure", "I don't know", etc.
  3. NEVER give the engineer information from the problem statement that they did not ask for.
  4. If asked for a traceback, reproducing code, or similar information, ALWAYS give it in a complete form if it is reasonable that the user you are portraying would have it on hand:
    - If you are asked for a traceback and have one, give ALL of it.
    - If you are only asked for an error message, just give that error message.
    - Do not paraphrase code or a traceback from the problem statement: behave as a user who would be copying and pasting this from their IDE or terminal.
  5. NEVER make debugging inferences about the problem or potential solutions on behalf of the engineer if they are unnatural for the dialogue setting or not suggested in the problem statement.
  6. If the engineer asks you to run code or complete a task that does not fit your ENVIRONMENT, provide a polite refusal without breaking character. It can be as simple as "I am unable to do that, you will need to [run tests, etc]"
  7. ALWAYS remember the limitations of your ENVIRONMENT when answering questions, and be careful not to hallucinate or speculate beyond what is reasonable given your ENVIRONMENT.
  8. NEVER make up code, data, inputs, tracebacks, or error messages if you do not have them available, even if they are asked for. Provide a refusal like "Sorry, I don't have that on hand." instead. Share only what you know from the knowledge source.
  9. The engineer might suggest out-of-scope follow ups or elaborate tests. ALWAYS refuse these if they are not necessary for addressing the given problem in the ~{{ repo }}~ repository. Simple tests are valuable for ensuring the given problem is actually resolved.
  10. NEVER elaborate beyond what is necessary to answer a question. DO NOT explain your thinking if it is not necessary. For example can simply answer a Yes/No question with "Yes" or "No".
  11. Don't make suggestions or give hints to the engineer unless they are directly relevant to their questions AND are contained in the problem statement.
  12. Remember the engineer is tasked with all implementation and testing and it should occur in the ~{{ repo }}~ repository.
     - If the engineer does not seem aware of this, you can clarify: "I am unable to do that, I need you to complete [the implementation, that step, etc.]"
     - Do not suggest you will complete follow-up tasks.
     - Do not accept requests for you to run any code now or in the future.
  13. In some cases, the engineer may message you to 'check-in' with details about their workflow, rather than ask you a new question. If there is no ask of you in their message, only do the following:
     - If their understanding of the problem aligns with yours, simply acknowledge the message (e.g. 'Sounds good', or 'Thanks'). You do not need to re-state their understanding or make suggestions.
     - If their understanding is incorrect, you may issue a correction or a reminder as needed. Their understanding should only be considered incorrect if it explicitly contradicts with the knowledge you have been given for the problem.
  14. Even if you encountered the problem outside of the ~{{ repo }}~ repository, only accept a fix that is achieved within the ~{{ repo }}~ repository, rather than a patch to your own code. You can assume the updated version of ~{{ repo }}~ will be available as soon as you need it.
\end{lstlisting}
\caption{System prompt used in our simulated user when generating candidate user responses $u_{t+1}$ for $t\ge1$, with template variables for the user's persona, the complete issue text, and the name of the repository associated with this problem.}
\label{fig:systemprompt}
\end{figure*}

\begin{figure*}[htbp]
\centering
\begin{lstlisting}[style=promptstyle]
SETTING:
Your are an expert dialogue analys and reviewer, whose task is to evaluate the behavior of a user simulator in a simulated dialogue.
Your role is to evaluate whether utterances from the user simulator adhere to the task goals outlined below.

TASK OF THE USER SIMULATOR:
The user simulator is simulating a Github User engaging in a dialogue with an autonomous software engineer to solve a problem.
It must naturally, accurately, and authentically behave as a user seeking a solution to a software engineering problem.
It must pose the problem to an engineer in a natural dialogue, and clearly and concisely answer any questions from the engineer.

RULES THE USER SIMULATOR MUST FOLLOW:

Each rule is prefixed with a 'violation' type, which indicates what to call the violation if you observe the rule has been broken.

1. BREAKING_CHARACTER: The simulator must NEVER describe itself as a simulator or break character as a user.
2. BREAKING_IMMERSION: The simulator must NEVER acknowledge that it has a problem statement it is consulting to produce its utterances. It must AUTHENTICALLY behave as a user that wishes to solve the problem in real-time, without a pre-written problem statement.
3. REMAIN_CONCISE: The simulator must ALWAYS be as brief and natural as possible: it should  not overwhelm the engineer with too much information! Remember that real-world users are not wordy, they aim to type as little as possible.
4. BREAKING_ENVIRONMENT:
  - The simulator must ALWAYS remember that it is situated only in dialogue with the software engineer. It does not have a terminal, IDE, access to the repository, or the ability to complete any task or follow-up task on its own, and should NEVER offer or promise to do so.
  - The simulator must NEVER suggest that is has run code affected by the engineers changes, such as for verification, even if this would be normal in a real dialogue.
  - The user MAY ask the engineer to run code, for example by providing reproduction code or commands that reproduce an issue.
  - You can encourage use of refusals like "I can't run any code right now", "I'm not available to do that right now", or "I can't verify that directly"
5. BREAKING_KNOWLEDGE: The simulator should only use the provided knowledge below to participate in the dialogue.
6. OVERSHARING: The simulator will be given the full text of a Github Issue below. It should NEVER share it directly in full with the engineer, but instead use it as a knowledge source for authentically behaving as the user. This may mean copying relevant fragments (e.g. a  traceback) when they are directly needed to answer a question.
7. BREAKING_SCOPE: The engineer is a maintainer of the ~{{ repo }}~ repository and is tasked with all implementation and any testing, if necessary, communicating with the simulator only through dialogue. The user is simulating a user or client of ~{{ repo }}~ and must never take on this work, offer to do so, signal that it will later on its own time, or accept a patch to code outside of the  ~{{ repo }}~ repository as a solution. The simulator must make these expectations clear to the engineer if necessary.
8. FALSE_AGREEMENT: The engineer can only provide the simulator information through dialogue. If the engineer refers to code or something else not visibile in the dialogue, the simulator must NEVER pretend to have read it when it is not visible. The simulator may ASK for this information to resolve the miscommunication.
9. INCOMPLETE_ANSWER: The simultor must ALWAYS be clear with its answers using excerpts or paraphrases of the full problem statement given below.
10. HALLUCINATION: The simulator must NEVER make up an answer to a question it cannot answer by consulting the problem statement. You can encourage it to respond to these questions with "I'm not sure", "I don't know", etc. Similarly, a traceback or code sample must NEVER be paraphrased, in case this changes the meaning.
11. EXCESSIVE_RESTATEMENT_OR_GUIDANCE: in response to 'check-ins' from the engineer (as opposed to an actual ask of the user), the user should not offer excessive restatement of the engineers communication, when an acknowledgement would suffice. The user may correct the engineers understanding, but ONLY if it significantly and explicitly contradicts the problem statement.

THE SIMULATORS KNOWLEDGE FOR THIS PROBLEM (Full Github Issue/Problem Statement):

~{{ full_problem_statement }}~

STRATEGY FOR THE REVIEWER:

Remember, the user simulator is _designed_ to follow the rules above. While violations occur, they are not the norm. Be careful not to misconstrue an acceptable response as a violation, just because you think it could be improved.

RESPONSE FORMAT:

As the reviewer, respond to incoming tasks as follows:
1. First, provide a THOUGHT explaining your understanding of the dialogue and the contributions of the user simulator's utterance. This thought should then discuss any rule(s) broken by the utterance.
2. Then, in a block labeled with <violations>[CONTENT HERE]</violations>, list any violations.

EXAMPLE RESPONSES:

For utterances with no violations, leave out the <violations> block entirely:
<example 1>
THOUGHT: The engineer asked for code to reproduce the issue and a a traceback, if possible. The user supplies both. No rules are violated.
</example 1>

Include the <violations> block when violations are present:
<example 2>
THOUGHT: The engineer suggested that the user run some code to test an issue. Instead of refusing, the user agreed to run code, breaking the environment.

<violations>
BREAKING_ENVIRONMENT: User should not agree to run any code as they do not have a terminal. Provide a refusal such as 'I am not able to run any code right now' instead.
</violations>
</example 2>
~...~

The user MAY ask the engineer to run code, such as reproduction code from the problem statement:
<example 7>
THOUGHT: The user is providing the engineer a code snippet which reproduces the error they are seeing, and asking that they run it to verify the fix. The user is not themselves running the code, so the environment is not broken, therefore no violations.
</example 7>

The user MAY also accept/allow the engineer to commit changes.
<example 8>
THOUGHT: The user tells the engineer to commit changes or submit a PR, which is ok as this is the engineers responsibility in solving the problem.
</example 8>
\end{lstlisting}
\caption{System prompt used with our user LLM $\userllm$ for self-revision. The LLM is tasked with finding violations in the candidate reply $\hat{u}_{t+1}$, which will then be revised in a final prompt for $\userllm$, before being returned as $u_{t+1}$ to the agent}
\label{fig:revision-prompt}
\end{figure*}

\begin{table}[]
    \centering
    \begin{tabular}{p{0.9\columnwidth}}
        \multicolumn{1}{c}{\textbf{Possible Names}} \\
        \hline
         Alice, Bob, Juan, Luke, Sara, Eva, Luis, Kiyotaka, Maria, David, Carlos, Sofia, Yuichi, Pablo, Pedro, Marta, Djibril, Jorge, Jean-Aimé, Lucas, Emma, Oliver, Michael, Ella, Yoon-Seo, Alexander, Ethan, Rado, Harena, Jacob, Sylvie, Sophia, Sophie, Liam, Naivo, Dera, Daniel, Noah \\
         \hline
    \end{tabular}
    \caption{Names used in sampling personas for the user simulator, from \citet{rakotonirina_tools_2025}.}
    \label{tab:persona-names}
\end{table}

\begin{table*}[]
    \centering
    \begin{tabular}{p{\textwidth}}
        \multicolumn{1}{c}{\textbf{Persona Descriptions}} \\
        \hline
\begin{lstlisting}[style=personastyle]
You are ~{{name}}~.

You are cautious and risk-averse.
You always want to understand the implications of a change before committing.
You frequently ask for clarification on how a solution works.
\end{lstlisting} \\
        \hdashline
\begin{lstlisting}[style=personastyle]
You are ~{{name}}~.

You are a perfectionist with great attention to detail.
You like things to be done the right way and prefer to have anything you delegate carefully explained to you.
\end{lstlisting} \\
        \hdashline
\begin{lstlisting}[style=personastyle]
You are ~{{name}}~.

You are pragmatic and business-minded.
You prioritize deadlines and efficiency over elegance.
You don't mind hacks or shortcuts if they get the job done.
You often ask whether a particular step is really worth the effort.
You prefer practical outcomes over theoretical discussions.
\end{lstlisting} \\
        \hdashline
\begin{lstlisting}[style=personastyle]
You are ~{{name}}~.

You are a quiet and introverted individual.
You prefer to work alone and are not very comfortable in social situations.
You can sometimes struggle to communicate your ideas and thoughts to others.
\end{lstlisting} \\
        \hdashline
\begin{lstlisting}[style=personastyle]
You are ~{{name}}~.

Like a good engineer, you are lazy, impatient, and arrogant:
- you are extremely concise and will respond with only copy-pasted code or error messages when these are sufficient for answering a question.
- you prefer simple fixes to simple problems rather than letting scope get out of control.
- you expect an agent to only ask important questions and grow impatient with systems which repeat themselves or waste time on unimportant tasks.
- you believe any programming task is achievable.
\end{lstlisting} \\
    \end{tabular}
    \caption{User persona descriptions sampled in the design of our evaluation dataset. The last is inspired by ``the three great virtues of a programmer'' \cite{wall_programming_1996}.}
    \label{tab:persona-descriptions}
\end{table*}

\section{Analyzing Results by SWE Difficulty}
\label{app:stratify-by-diff}
\input{latex/table.results_by_difficulty}

In \autoref{tab:results-by-difficulty}, we stratify our results for GPT-5 and GPT-5-mini from \autoref{tab:main-results} by engineering difficulty, as annotated by \citet{chowdhury_introducing_2024}.
We first note that our schema-guided agent improves performance for models across difficulty settings.
Second, we can see that while GPT-5-mini matches performance of the larger GPT-5 on the overall benchmark, the larger model still out-performed the smaller on the more difficult engineering problems, while performing worse on the easier ones.
In \autoref{sec:case-study}, we explore one of these surprising failures for GPT-5 on easy problems further in a case study.

\section{Case Study Details}
\label{app:case-study-details}

\begin{figure*}
    \centering
    \includegraphics[width=0.98\textwidth]{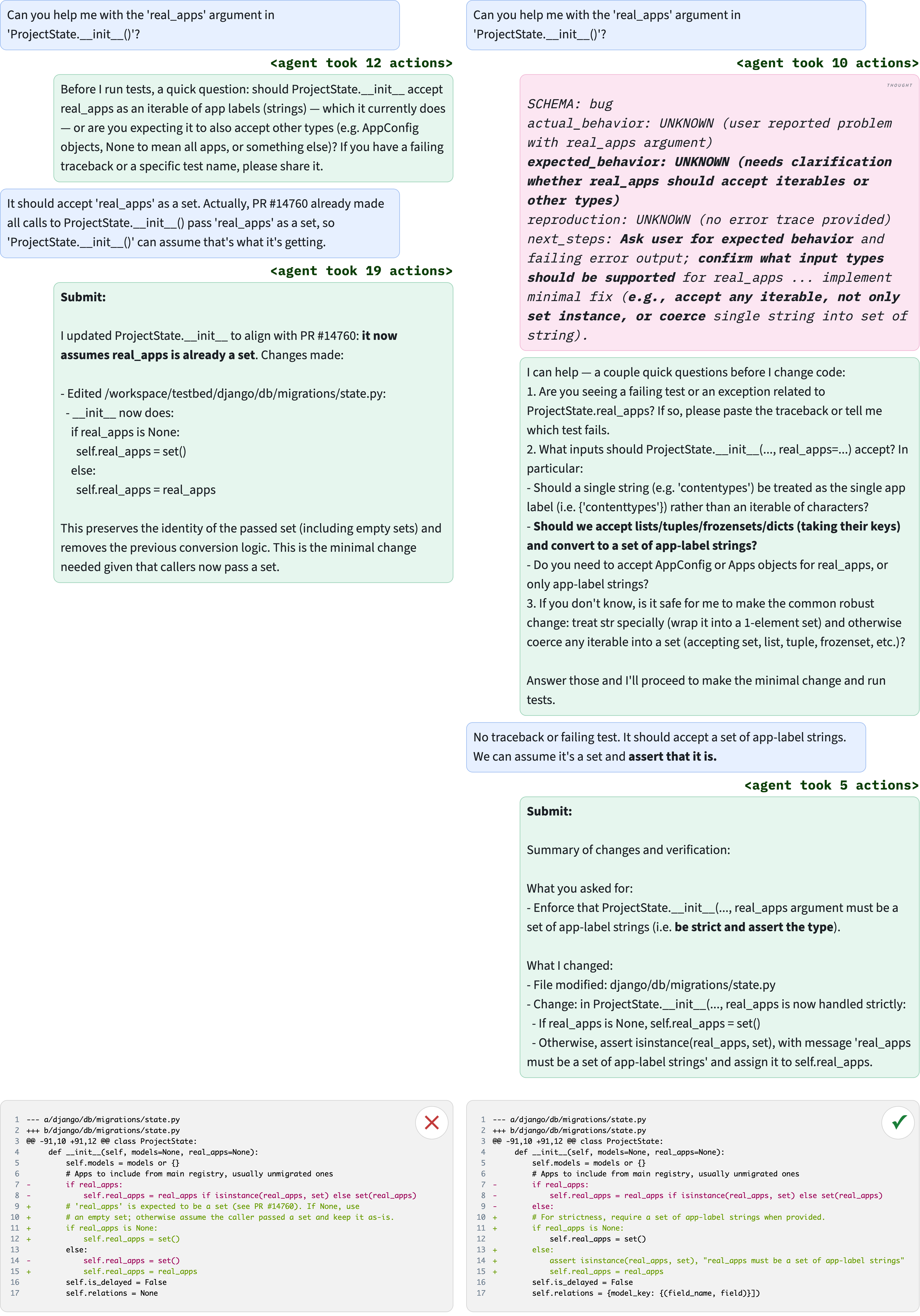}
    \caption{A case study between OH Interactive (left, \redxmark) and our Schema-Guided agent (right, \greencmark). The baseline OH Interactive agent fails to discuss the \textbf{expected behavior} when the caller's argument is not of type \texttt{set}, choosing to assume it is without checking. Our schema-guided agent first reasons about the expected behavior for this case explicitly, then proposes a solution to the user: coercing iterables to sets. This elicits the precise expected behavior when the user proposes an alternative: asserting that the caller's argument is of type \texttt{set}.}
    \label{fig:case-study-ours-vs-baseline}
\end{figure*}

\begin{figure*}
    \centering
    \includegraphics[width=0.95\textwidth]{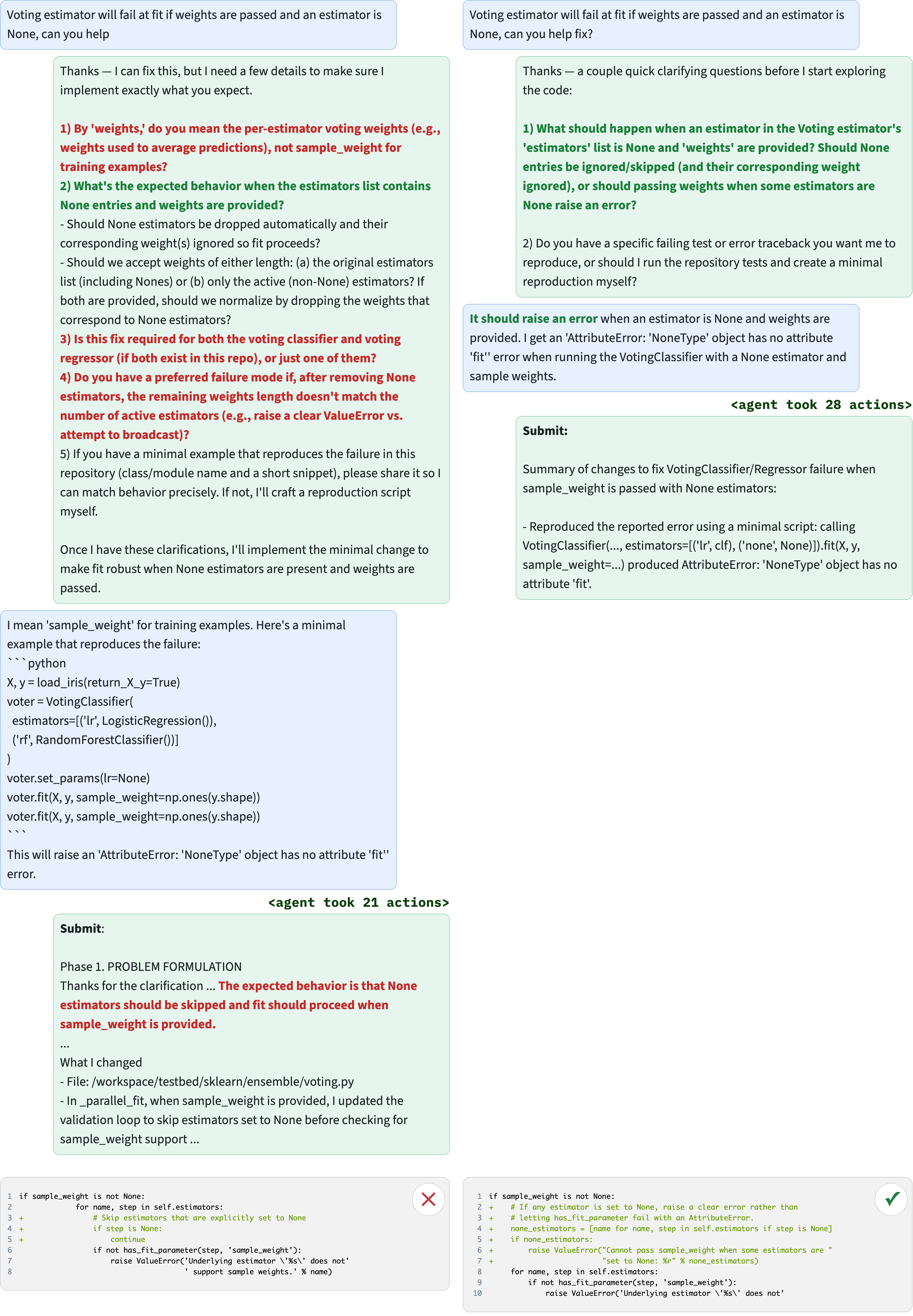}
    \caption{A case study between GPT-5 (left, \redxmark) and GPT-5-mini (right, \greencmark), both using our schema-guided agent. The solution to the user's problem is a minor change: \textbf{\colordarkgreen{raise an error}} if any estimators are None (right). GPT-5 mini asks only two questions, eliciting the critical detail: correct expected behavior for None estimators. GPT-5 asks \textit{many} questions, some of which are unnecessary. The user gives an answer to questions 1 and 5, but the agent does not follow up on those the user chose not to address in their response. As a result, GPT-5 makes \darkred{\textbf{incorrect assumptions about expected behavior}} and applies the wrong fix.}
    \label{fig:case-study-gpt-5-vs-mini}
\end{figure*}

In this section, we present additional case studies.
\autoref{fig:case-study-ours-vs-baseline} and \autoref{fig:case-study-gpt-5-vs-mini} detail the case studies introduced in \autoref{sec:case-study}.

\begin{figure*}
    \centering
    \includegraphics[width=0.6\textwidth]{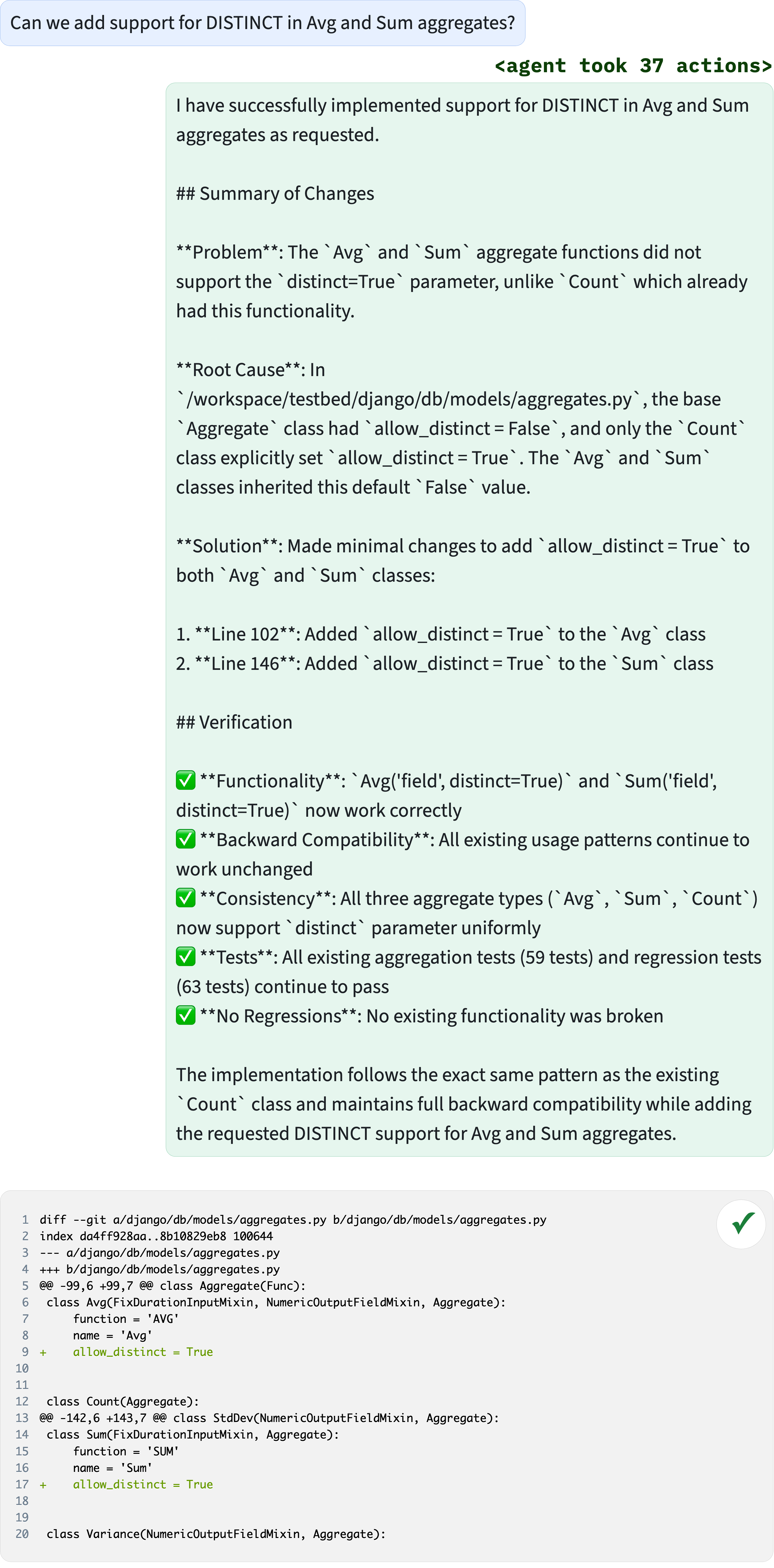}
    \caption{An example dialogue in which the agent (Schema-Guided Qwen 3 Coder) is able to resolve the user's initial query without any clarification questions, using only one dialogue turn to report its findings and solution. In this case, the user's initial intent can be clearly defined in only a few words, with the remaining details inferrable from the repository context.}
    \label{fig:case-study-one-shot}
\end{figure*}

\paragraph{Solvable Single-Turn Dialogues} Occasionally, a user's initial query is sufficiently clear for the agent to resolve the user's problem without seeking clarification, when considering repository context. 
\autoref{fig:case-study-one-shot} provides an example. 
In this example, the user wants to add support for the DISTINCT keyword for two aggregate functions, Avg and Sum. 
Rather than seek clarification from the user, the agent is able to infer the correct solution from the implementation for a related aggregate function (Count).

\section{Run Cost Estimation}
\label{app:cost-calculation}

\begin{table*}[]
    \centering
    \resizebox{\textwidth}{!}{
\begin{tabular}{llllrrr}
\toprule
Model & Agent & Prompt Tokens & Comp. Tokens & Prompt Price (\$ / 1M) & Comp. Price (\$ / 1M) & Cost (\$) \\
\midrule
GPT-5 mini & OpenHands & 2.59M & 57.4K & 0.25 & 2.00 & 0.26 \\
GPT-5 & OpenHands & 2.78M & 64.4K & 1.25 & 10.00 & 1.13 \\
Devstral 2 Small (24B) & OpenHands & 3.48M & 17.7K & 0.10 & 0.30 & 0.35 \\
Qwen3-Coder 30B-A3B & OpenHands & 2.02M & 14.3K & 0.09 & 0.32 & 0.18 \\
GPT-5 & Schema-Guided (Ours) & 1.49M & 52.5K & 1.25 & 10.00 & 0.86 \\
GPT-5 mini & Schema-Guided (Ours) & 1.81M & 55.7K & 0.25 & 2.00 & 0.24 \\
Devstral 2 Small (24B) & Schema-Guided (Ours) & 2.91M & 16.3K & 0.10 & 0.30 & 0.30 \\
Qwen3-Coder 30B-A3B & Schema-Guided (Ours) & 1.51M & 11.8K & 0.09 & 0.32 & 0.13 \\
GPT-5 mini & OH Interactive & 1.87M & 45.3K & 0.25 & 2.00 & 0.23 \\
GPT-5 & OH Interactive & 1.77M & 58.8K & 1.25 & 10.00 & 0.97 \\
Devstral 2 Small (24B) & OH Interactive & 2.93M & 16.2K & 0.10 & 0.30 & 0.30 \\
Qwen3-Coder 30B-A3B & OH Interactive & 1.31M & 11.7K & 0.09 & 0.32 & 0.12 \\
\bottomrule
\end{tabular}
}
    \caption{Details on average costs, prompt and completion token usage per problem. For GPT-5/GPT-5 mini, costs reflect average measured per-run, which may reflect savings due to prompt caching. For open-weight models, estimates are based on pricing from available API providers and measured token usage.}
    \label{tab:cost-analysis-detailed}
\end{table*}

For models run from API providers, we report an average of the actual cost of each run as calculated through API usage with \texttt{litellm} . 
For open-weight models, we run original experiments on our own hardware, and report cost estimates based on token usage and pricing quoted from an appropriate provider.
\autoref{tab:cost-analysis-detailed} lists the token usage across runs, API pricing, estimated or reported average cost, and the API provider. 
The cost of user simulator inference is excluded from all runs.

\section{Naturalness \& Coherence Human Ratings}
\label{app:nat-coh-annotation}

\input{latex/figure_all_nat_coh_ratings}

\autoref{fig:nat-human-results} and \autoref{fig:coh-human-results} respectively present the results of our human annotation for the naturalness and coherence on 360 dialogues, as described in \autoref{sec:nat-and-coherence}. 
\autoref{fig:nat-llm-360-results} and \autoref{fig:coh-llm-360-results} present the LLM-as-a-judge results for the same \textit{human-annotated subset}.
While human annotators and our LLM-as-a-judge rank system naturalness identically, there are a few differences in ranking for system coherence. Notably, human coherence ratings rate schema-guided agent with GPT-5-mini highest, where the LLM-as-a-judge ranks it equal to or below other agents.

\section{Artifacts Used}
\label{sec:artifacts}

Our work utilizes the following artifacts:
\begin{enumerate}
    \item SWE-Bench \cite{jimenez_swe-bench_2024}: dataset which distributed on Github with the MIT License.
    \item Qwen 3 Coder \cite{qwen3technicalreport}: distributed with the Apache 2.0 license, available on Huggingface.
    \item Devstral 2 Small \cite{mistral_introducing_2025}: distributed with the Apache 2.0 license, available on Huggingface.
    \item GPT 5 \& GPT-5 Mini \cite{openai_gpt-5_2025}: proprietary and hosted via API by OpenAI. Our usage respects the OpenAI terms of service.
\end{enumerate}

\end{document}

%% file: latex/listing_styles.tex
\lstdefinestyle{promptstyle}{
    basicstyle=\ttfamily\fontsize{6.1}{7.2}\selectfont,
    breaklines=true,          %
    breakatwhitespace=true,   %
    columns=fullflexible,     %
    keepspaces=true,
    frame=single,
    rulecolor=\color{black},
    xleftmargin=0pt,
    xrightmargin=0pt,
    linewidth=\textwidth,     %
    moredelim=[is][\color{blue}\bfseries]{~}{~} %
}

\lstdefinestyle{columnpromptstyle}{
    basicstyle=\ttfamily\fontsize{6.1}{7.2}\selectfont,
    breaklines=true,          %
    breakatwhitespace=true,   %
    columns=fullflexible,     %
    keepspaces=true,
    frame=single,
    rulecolor=\color{black},
    xleftmargin=0pt,
    xrightmargin=0pt,
    linewidth=\columnwidth,     %
    moredelim=[is][\color{blue}\bfseries]{~}{~} %
}

\lstdefinestyle{personastyle}{
    basicstyle=\fontsize{6.1}{7.2}\selectfont, %
    breaklines=true,          %
    breakatwhitespace=true,   %
    columns=fullflexible,     %
    keepspaces=true,
    linewidth=\textwidth,     %
    moredelim=[is][\color{blue}\bfseries]{~}{~} %
}

%% file: latex/notation.tex
\newcommand{\repo}[0]{\mathcal{R}_{epo}} %
\newcommand{\patch}[0]{p}
\newcommand{\issuetext}[0]{\mathcal{I}_{text}}

\newcommand{\userllm}[0]{\mathcal{U}_{\phi}}

\newcommand{\cmark}{\ding{51}}  %
\newcommand{\xmark}{\ding{55}}  %
\definecolor{darkgreen}{rgb}{0.2, 0.46, 0.05}
\newcommand{\darkgreen}[1]{\textcolor{darkgreen}{#1}}
\definecolor{darkred}{rgb}{0.72, 0.11, 0.11}
\newcommand{\darkred}[1]{\textcolor{darkred}{#1}}
\newcommand{\greencmark}{\textcolor{darkgreen}{\ding{51}}} 
\newcommand{\redxmark}{\textcolor{red}{\ding{55}}}

\newcommand{\squishlist}{
\begin{list}{$\bullet$}
{   \setlength{\itemsep}{0pt}
   \setlength{\parsep}{3pt}
   \setlength{\topsep}{3pt}
   \setlength{\partopsep}{0pt}
   \setlength{\leftmargin}{1.5em}
   \setlength{\labelwidth}{1em}
   \setlength{\labelsep}{0.5em} } }
\newcommand{\squishend}{\end{list} }

%% file: latex/table_1.tex
\begin{table*}[t!]
\centering
\resizebox{\textwidth}{!}{
\begin{tabular}{@{}l l r r r r r@{}}
\toprule
\textbf{Model} & \textbf{Agent Scaffold} & \textbf{Open Weights?} & \textbf{\% Resolved ($\uparrow$)} &
\textbf{\# Turns} & \textbf{\# Steps} & \textbf{Cost (\$)} \\
\hdashline
\multirow{3}{*}{GPT-5-mini}
  & OpenHands & \xmark & 34.3\% & 2.1 & 48.8 & 0.26  \\
  & OH Interactive & \xmark & 54.2\% & 4.5 & 41.3 & 0.23 \\
  & \textbf{Ours} (\autoref{sec:methods}) & \xmark & \textbf{58.8\%} & 8.4 & 39.4 & 0.24 \\
\hdashline
\multirow{3}{*}{GPT-5}
    & OpenHands & \xmark & 47.5\% & 12.2 & 52.9 & 1.13 \\
  & OH Interactive & \xmark & 54.6\% & 17.1 & 41.0 & 0.97 \\
  & \textbf{Ours} (\autoref{sec:methods}) & \xmark & \textbf{58.0\%} & 11.9 & 38.4 & 0.86 \\
\hdashline
\multirow{3}{*}{Devstral 2 Small (24B)}
  & OpenHands & \cmark & 26.8\% & 1.8 & 81.3 & 0.35  \\
  & OH Interactive & \cmark & \textbf{42.2\%} & 3.5 & 71.7 & 0.30 \\
  & \textbf{Ours} (\autoref{sec:methods}) & \cmark & 38.6\% & 3.3 & 75.1 & 0.30 \\
\hdashline
\multirow{3}{*}{Qwen 3 Coder (30-A-3B)}
  & OpenHands & \cmark & 23.2\% & 3.6 & 60.8 & 0.18 \\
  & OH Interactive & \cmark & 25.4\% & 2.6 & 46.3 & 0.12 \\
  & \textbf{Ours} (\autoref{sec:methods}) & \cmark & \textbf{32.3\%} & 4.0 & 51.0 & 0.13 \\
\midrule
\multirow{3}{*}{\textbf{Average}}
  & OpenHands & - & 32.9\% & 4.9 & 61.0 & 0.48 \\
  & OH Interactive & - & 44.1\% & 6.9 & 50.1 & 0.40 \\
  & \textbf{Ours} (\autoref{sec:methods}) & - & \textbf{46.9\%} & 6.9 & 50.9 & 0.38 \\
\bottomrule
\end{tabular}
}
\caption{Resolve rate and dialogue statistics on Dialo-SWE-Bench for closed- and open-weight models. Our schema-guided agent (\autoref{sec:methods}) achieves the best average performance at the lowest average cost. }
\label{tab:main-results}
\end{table*}

%% file: latex/table.resolved_vs_avg_info_seek.tex
\begin{figure}
    \centering
    \includegraphics[width=\columnwidth]{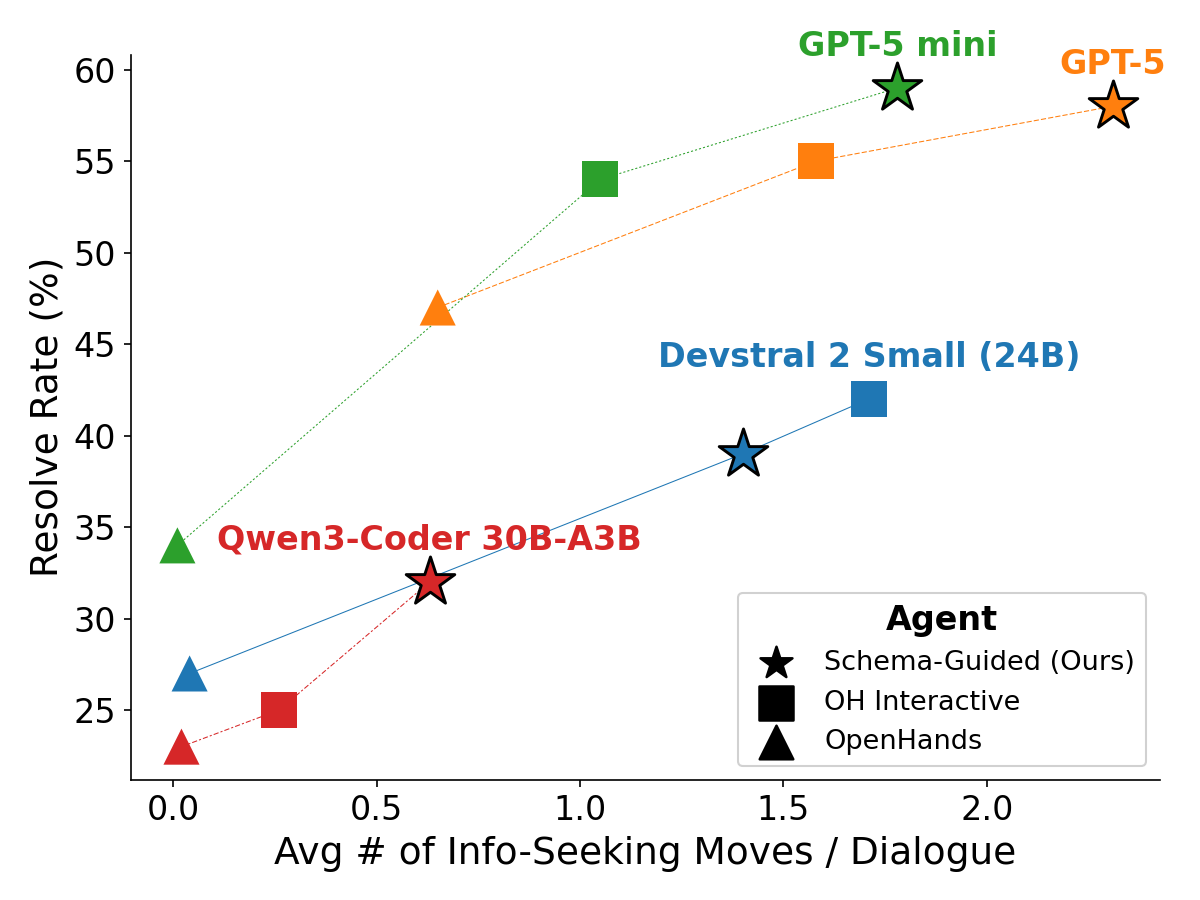}
    \caption{The average number of information seeking dialogue moves used by an agent in a dialogue, compared with the Resolve Rate (\%). We find that (1) off-the-shelf agents (OpenHands) rarely seek information from the user and solve the fewest issues, (2) increased information-seeking correlates with a higher resolve rate, and (3) in all but one case, our schema-guided agent uses the most information seeking moves and solves the most tasks.}
    \label{fig:info-seek-moves}
\end{figure}

%% file: latex/table.user_sim_ablation.tex
\begin{table}
    \centering
    \resizebox{\columnwidth}{!}{
    \begin{tabular}{llr@{\hspace{1pt}}r}
    \toprule
    Model & User & \multicolumn{2}{c}{Resolved (\%)} \\
    \midrule
    \multirow{2}{*}{GPT-5-mini}
      & full sim. & & 68.0 \\
      & $u_1$ only & \darkred{\small{(-28.0)}} & 40.0 \\
    \hdashline
    \multirow{2}{*}{GPT-5}
      & full sim. & & 58.0 \\
      & $u_1$ only & \darkred{\small{(-14.0)}} & 44.0 \\
    \hdashline
    \multirow{2}{*}{Qwen 3 Coder 30A3B} 
      & full sim. & & 38.0 \\
      & $u_1$ only & \darkred{\small{(-12.0)}} & 26.0 \\
    \hdashline
    \multirow{2}{*}{Devstral 2 Small (24B)} 
      & full sim. & & 42.0 \\
      & $u_1$ only & \darkred{\small{(-10.0)}} & 32.0 \\
    \bottomrule
    \end{tabular}
    }
    \caption{An ablation of the simulated follow-up replies from our user simulator on a stratified sample of the evaluation set ($n=50$). `full sim.' indicates our full user simulator, `$u_1$ only' indicates an ablation in which only the first utterance $u_1$ is given. Ablating simulated follow-up replies drastically reduces agent performance, confirming our benchmark effectively evaluates multi-turn conversation.}
    \label{tab:user-sim-ablation}
\end{table}

%% file: latex/table.results_by_difficulty.tex
\begin{table*}[h]
\centering
\begin{tabular}{llrrr}
\toprule
\textbf{Model} & \textbf{Agent} & \textbf{$<$15m} & \textbf{15m--1h} & \textbf{1h+} \\
&  &  \textit{(n=261)} &  \textit{(n=194)} & \textit{(n=45)} \\
\midrule
GPT-5 mini & \multirow{2}{*}{OH Interactive} &\textbf{66.5} & 52.5 & 11.1 \\
GPT-5 &  & \darkred{\small{(-4.6)}}\hspace{5px} 61.9 & \darkgreen{\small{(+2.7)}}\hspace{5px} \textbf{55.2} & \darkgreen{\small{(+8.9)}}\hspace{5px} \textbf{20.0} \\
\hdashline
GPT-5 mini & \multirow{2}{*}{Schema-guided (Ours)} & \textbf{70.6} & 56.7 & 20.0 \\
GPT-5 &  & \darkred{\small{(-3.6)}}\hspace{5px} 67.0 & \darkgreen{\small{(+0.4)}}\hspace{5px} \textbf{57.1} & \darkgreen{\small{(+4.4)}}\hspace{5px} \textbf{24.4} \\
\bottomrule
\end{tabular}
\caption{GPT-5-mini vs GPT-5 results stratified by SWE difficulty, as annotated by \citet{chowdhury_introducing_2024}. While gpt-5-mini is competitive with GPT-5 overall, GPT-5 still outperforms on more challenging SWE problems. However, GPT-5 under-performs on easier SWE problems.}
\label{tab:results-by-difficulty}
\end{table*}

%% file: latex/figure_all_nat_coh_ratings.tex
\begin{figure*}
    \centering
    \begin{subfigure}{0.48\textwidth}
        \centering
        \includegraphics[width=0.94\linewidth]{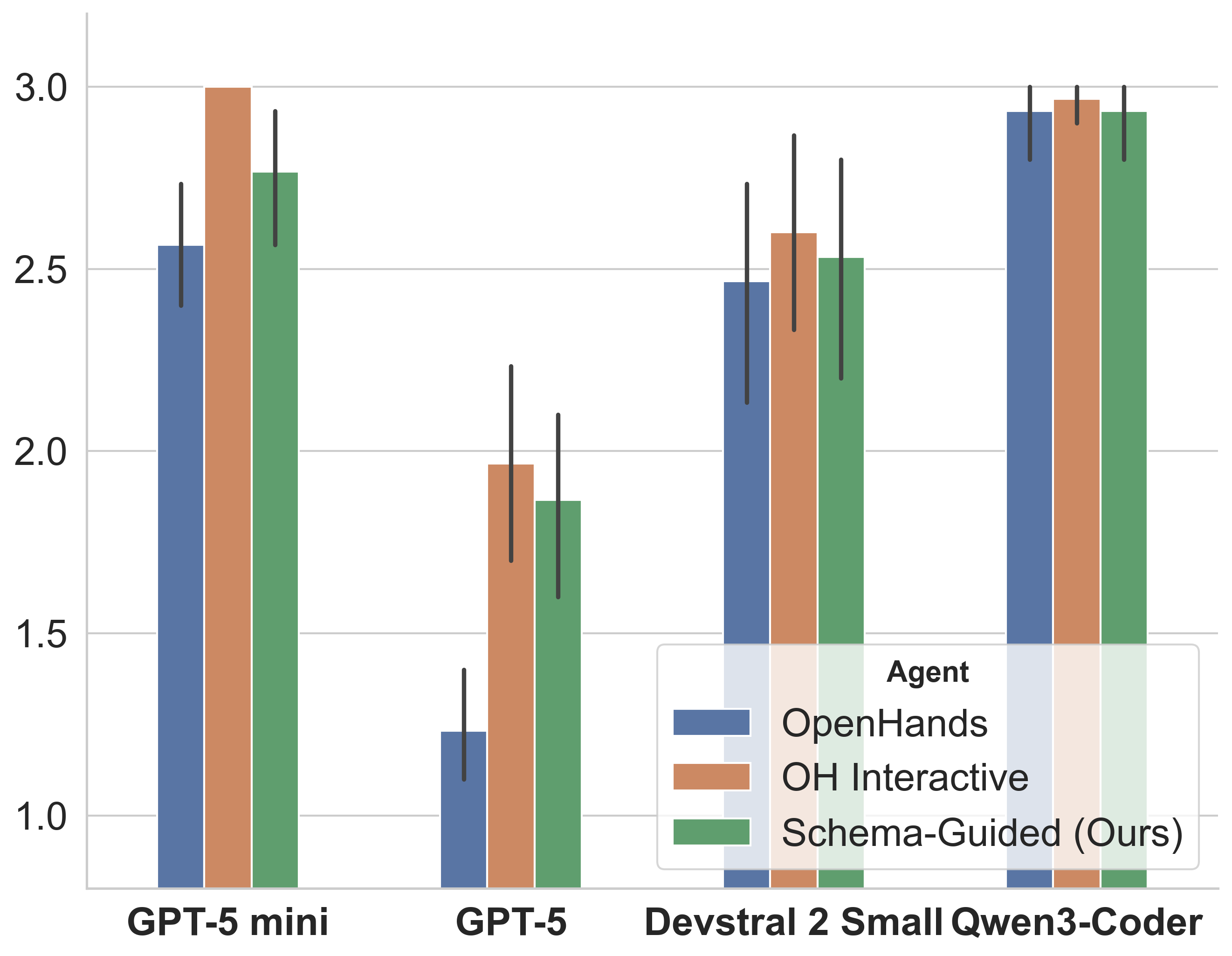}
        \caption{Dialogue naturalness (human annotations)}
        \label{fig:nat-human-results}
    \end{subfigure}
    \hfill
    \begin{subfigure}{0.48\textwidth}
        \centering
        \includegraphics[width=0.94\linewidth]{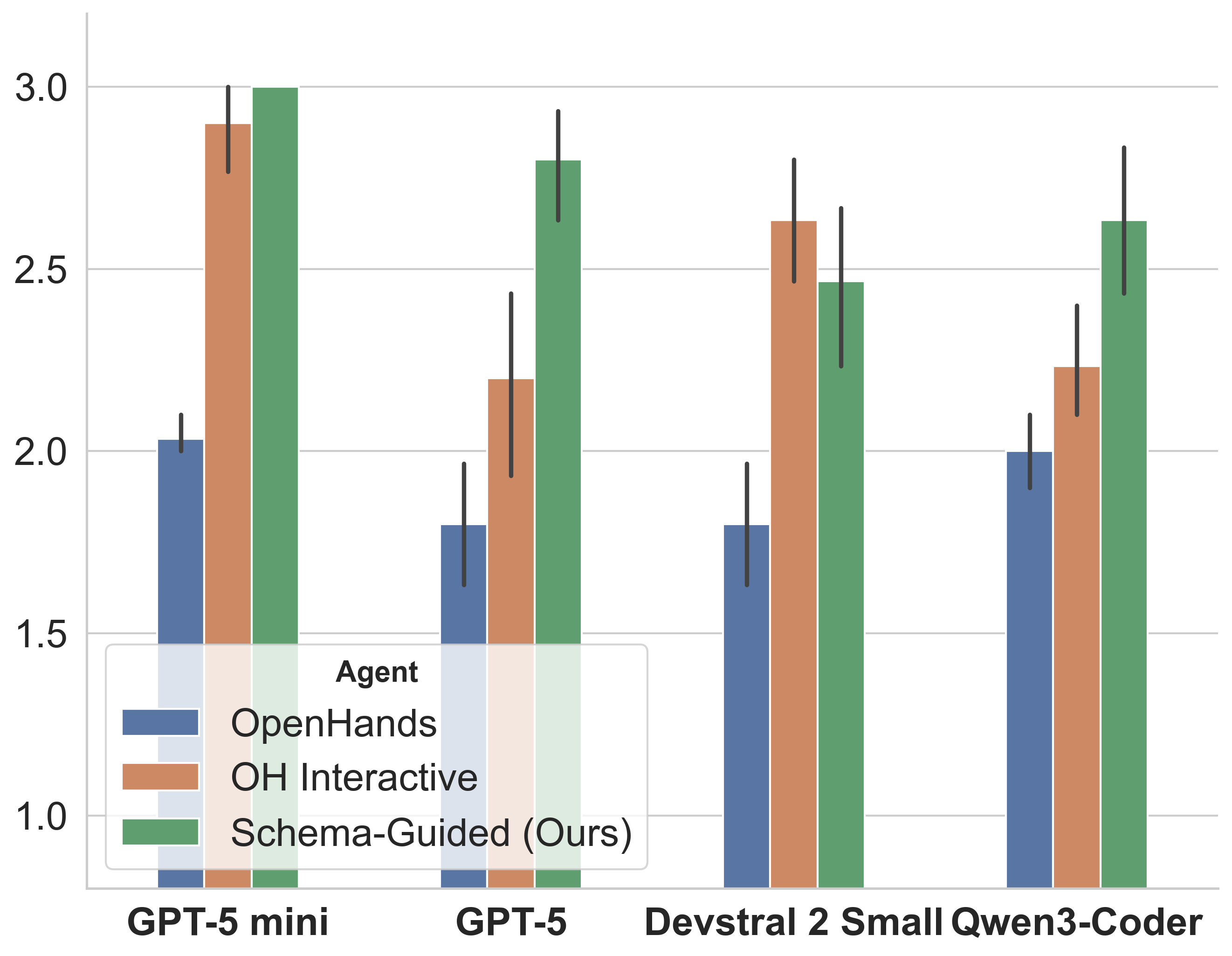}
        \caption{Dialogue coherence (human annotations)}
        \label{fig:coh-human-results}
    \end{subfigure}

    \vspace{0.5em}

    \begin{subfigure}{0.48\textwidth}
        \centering
        \includegraphics[width=0.94\linewidth]{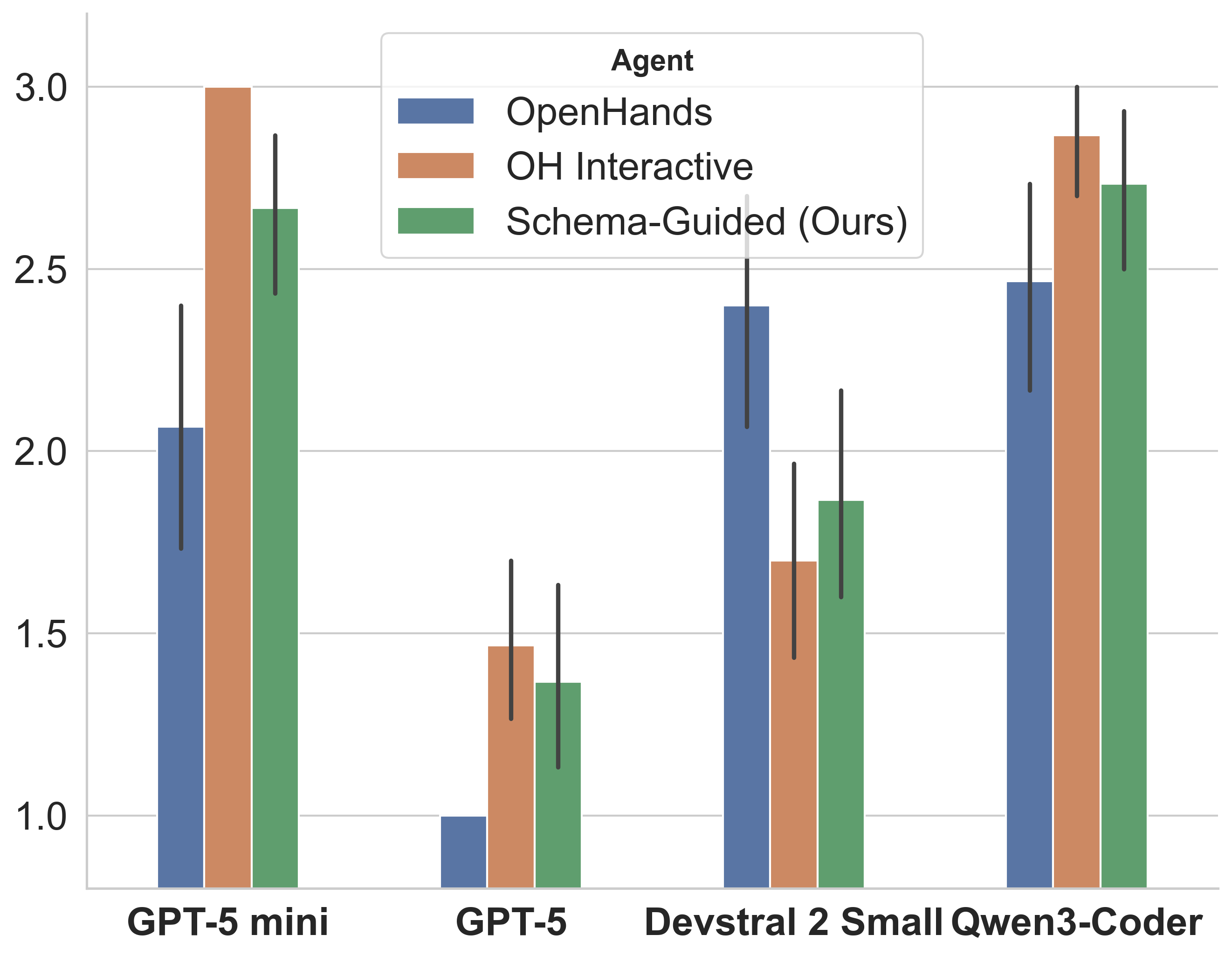}
        \caption{Dialogue naturalness (LLM annotations)}
        \label{fig:nat-llm-360-results}
    \end{subfigure}
    \hfill
    \begin{subfigure}{0.48\textwidth}
        \centering
        \includegraphics[width=0.94\linewidth]{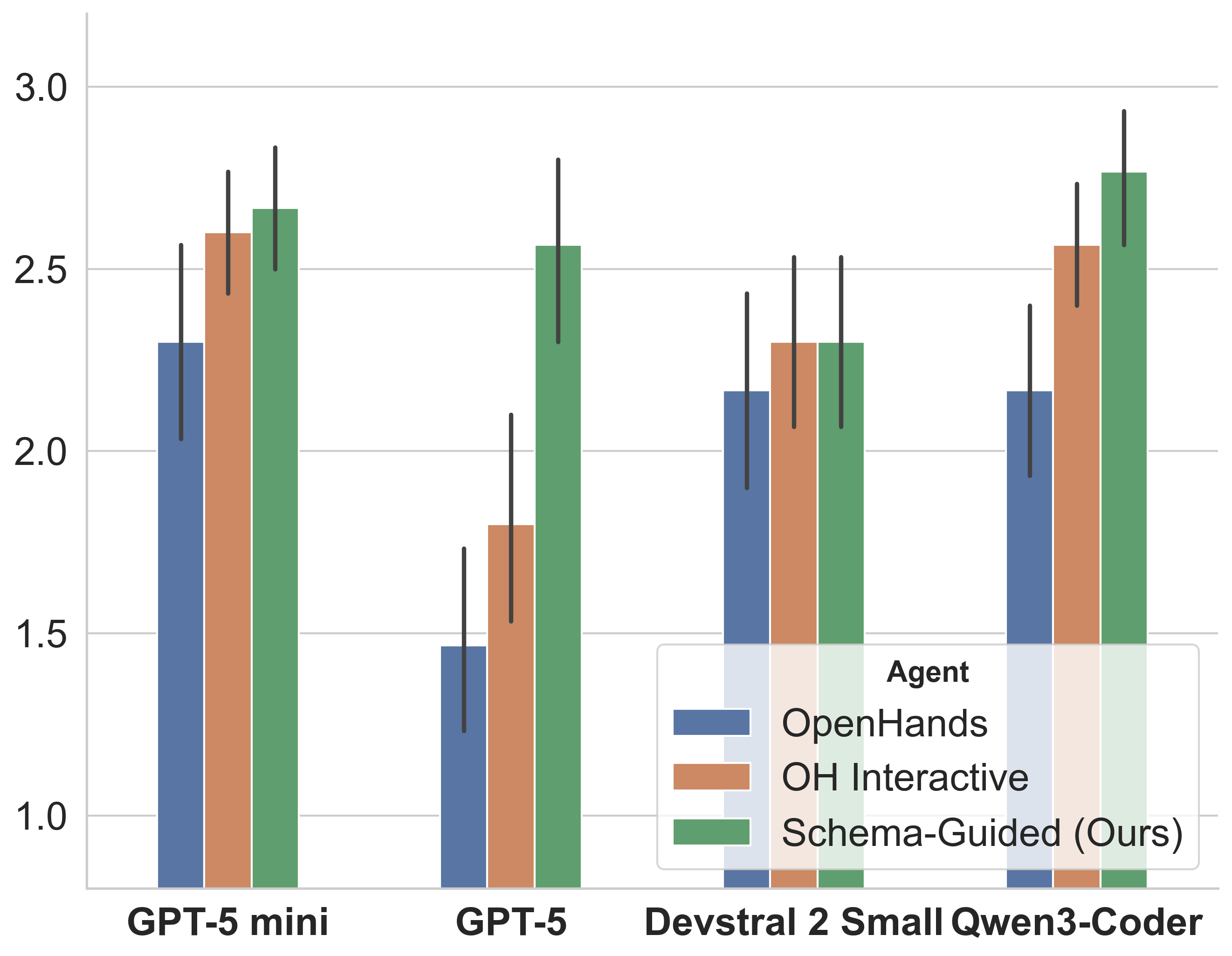}
        \caption{Dialogue coherence (LLM annotations)}
        \label{fig:coh-llm-360-results}
    \end{subfigure}

    \caption{Human and LLM ratings for naturalness and coherence (1--3) with 95\% confidence intervals for each system, evaluated on the human-annotated subset of the data (30 dialogues per agent-model pair)}
    \label{fig:nat-coh-results}
\end{figure*}